\documentclass[5p]{elsarticle}
\journal{Neural Networks}
\biboptions{authoryear}

\usepackage{amsmath,amssymb,amsfonts}
\usepackage{array}
\usepackage{bm}
\usepackage[monochrome]{color}
\usepackage{multirow}
\usepackage{hyperref}
\usepackage[switch]{lineno}\modulolinenumbers[5]

\newcolumntype{P}[1]{>{\centering\arraybackslash}p{#1}}

\newcommand{\Real}{{\mathbb{R}}}

\newcommand{\IE}{\textit{i}.\textit{e}., }
\newcommand{\EG}{\textit{e}.\textit{g}., }

\newcommand{\SecLab}[1]{\label{sec:#1}}
\newcommand{\SecRef}[1]{Section~\ref{sec:#1}}
\newcommand{\SecRefBare}[1]{\ref{sec:#1}}
\newcommand{\FigLab}[1]{\label{fig:#1}}
\newcommand{\FigRef}[1]{Fig.~\ref{fig:#1}}
\newcommand{\FigRefBare}[1]{\ref{fig:#1}}
\newcommand{\TblLab}[1]{\label{tbl:#1}}
\newcommand{\TblRef}[1]{Table~\ref{tbl:#1}}
\newcommand{\TblRefBare}[1]{\ref{tbl:#1}}
\newcommand{\EqLab}[1]{\label{eq:#1}}
\newcommand{\EqRef}[1]{Eq.~(\ref{eq:#1})}
\newcommand{\EqRefBare}[1]{(\ref{eq:#1})}

\newcommand{\MlapGraphs}{{\mathcal{G}}}
\newcommand{\MlapGraph}{{G}}
\newcommand{\MlapGraphEmb}[1]{{\bm{h}_{#1}}}
\newcommand{\MlapLayerEmb}[2]{{\bm{h}_{#1}^{(#2)}}}

\newcommand{\MlapNodes}{{\mathcal{N}}}
\newcommand{\MlapNode}{{n}}
\newcommand{\MlapNodePrime}{{n'}}
\newcommand{\MlapNodeSrc}{{\MlapNode_\mathrm{src}}}
\newcommand{\MlapNodeDst}{{\MlapNode_\mathrm{dst}}}
\newcommand{\MlapNodeFeat}[1]{{\bm{p}_{#1}}}
\newcommand{\MlapNodeEmbBare}[1]{{\bm{h}_{#1}}}
\newcommand{\MlapNodeEmb}[2]{{\bm{h}_{#1}^{({#2})}}}

\newcommand{\MlapEdges}{{\mathcal{E}}}
\newcommand{\MlapEdge}[2]{{e_{#1,#2}}}
\newcommand{\MlapEdgeFeat}[2]{{\bm{q}_{#1,#2}}}

\newcommand{\MlapNumLayer}{{L}}
\newcommand{\MlapLayer}{{l}}

\newcommand{\MlapNeighbor}[1]{{\mathrm{NBR}\left(#1\right)}}

\newcommand{\MlapLabels}{{\mathcal{Y}}}
\newcommand{\MlapLabel}{{y}}

\newcommand{\MlapMsg}[2]{{\bm{m}_{#1}^{(#2)}}}
\newcommand{\MlapMsgFn}[1]{{f_\mathrm{msg}^{(#1)}}}
\newcommand{\MlapColFn}[1]{{f_\mathrm{col}^{(#1)}}}
\newcommand{\MlapUpdFn}[1]{{f_\mathrm{upd}^{(#1)}}}
\newcommand{\MlapAggFn}{{f_\mathrm{agg}}}
\newcommand{\MlapGateFn}{{f_\mathrm{gate}}}
\newcommand{\MlapLayerGateFn}[1]{{f_\mathrm{gate}^{(#1)}}}
\newcommand{\MlapEdgeFn}[1]{{f_\mathrm{edge}^{(#1)}}}
\newcommand{\MlapGinNnFn}[1]{{f_\mathrm{NN}^{(#1)}}}
\newcommand{\MlapGinEps}[1]{{\epsilon^{(#1)}}}
\newcommand{\MlapJK}{{f_\mathrm{JK}}}

\newcommand{\MlapPool}{{\mathrm{Pool}}}
\newcommand{\MlapLayerPool}[1]{{\mathrm{Pool}^{(#1)}}}

\newcommand{\MlapWeight}[1]{{w^{(#1)}}}

\newcommand{\MlapEmbDim}{{d}}

\newcommand{\MlapMolhivOutWeight}{{\bm w_\mathrm{prob}}}
\newcommand{\MlapPPAClassEmbMat}{{\bm E}}
\newcommand{\MlapPPAClassEmb}[1]{{\MlapPPAClassEmbMat_{#1}}}

\makeatletter
\newcommand{\FigCap}[1]{\def\@captype{figure}\caption{#1}}
\newcommand{\TblCap}[1]{\def\@captype{table}\caption{#1}}
\makeatother

\begin{document}

\begin{frontmatter}

\title{Multi-Level Attention Pooling for Graph Neural Networks:\\Unifying Graph Representations with Multiple Localities}

\author{Takeshi D. Itoh}
\ead{itoh.takeshi.ik4@is.naist.jp}

\author{Takatomi Kubo\corref{cor}}
\cortext[cor]{Corresponding author.}
\ead{takatomi-k@is.naist.jp}

\author{Kazushi Ikeda}
\ead{kazushi@is.naist.jp}

\address{Division of Information Science, Graduate School of Science and Technology, Nara Institute of Science and Technology, 8916-5 Takayama-Cho, Ikoma, Nara 630-0192, Japan}

\begin{abstract}
Graph neural networks (GNNs) have been widely used to learn vector representation of graph-structured data and achieved better task performance than conventional methods.
The foundation of GNNs is the message passing procedure, which propagates the information in a node to its neighbors.
Since this procedure proceeds one step per layer, the range of the information propagation among nodes is small in the lower layers, and it expands toward the higher layers.
Therefore, a GNN model has to be deep enough to capture global structural information in a graph.
On the other hand, it is known that deep GNN models suffer from performance degradation because they lose nodes' local information, which would be essential for good model performance, through many message passing steps.
In this study, we propose multi-level attention pooling (MLAP) for graph-level classification tasks, which can adapt to both local and global structural information in a graph.
It has an attention pooling layer for each message passing step and computes the final graph representation by unifying the layer-wise graph representations.
The MLAP architecture allows models to utilize the structural information of graphs with multiple levels of localities because it preserves layer-wise information before losing them due to oversmoothing.
Results of our experiments show that the MLAP architecture improves the graph classification performance compared to the baseline architectures.
In addition, analyses on the layer-wise graph representations suggest that aggregating information from multiple levels of localities indeed has the potential to improve the discriminability of learned graph representations.

\end{abstract}


\end{frontmatter}

\tolerance 2000

\section{Introduction\SecLab{introduction}}

Graph-structured data are found in many fields.
A wide variety of natural and artificial objects can be expressed with graphs, such as molecular structural formula, biochemical reaction pathways, brain connection networks, social networks, and abstract syntax trees of computer programs.
Because of this ubiquity, machine learning methods on graphs have been actively studied.
Thanks to rich information underlying the structure, graph machine learning techniques have shown remarkable performances in various tasks.
For example, the PageRank algorithm \citep{page1999pagerank} computes the importance of each node in a directed graph based on the number of inbound edges to the node.
\citet{shervashidze2011weisfeiler} used a graph kernel method \citep{kondor2002diffusion} to predict chemical molecules' toxicity as a graph classification task.
Despite these promising applications, classical machine learning techniques on graphs require difficult and costly processes for manually designing features or kernel functions.

\begin{figure*}[t]
    \centering
    \includegraphics{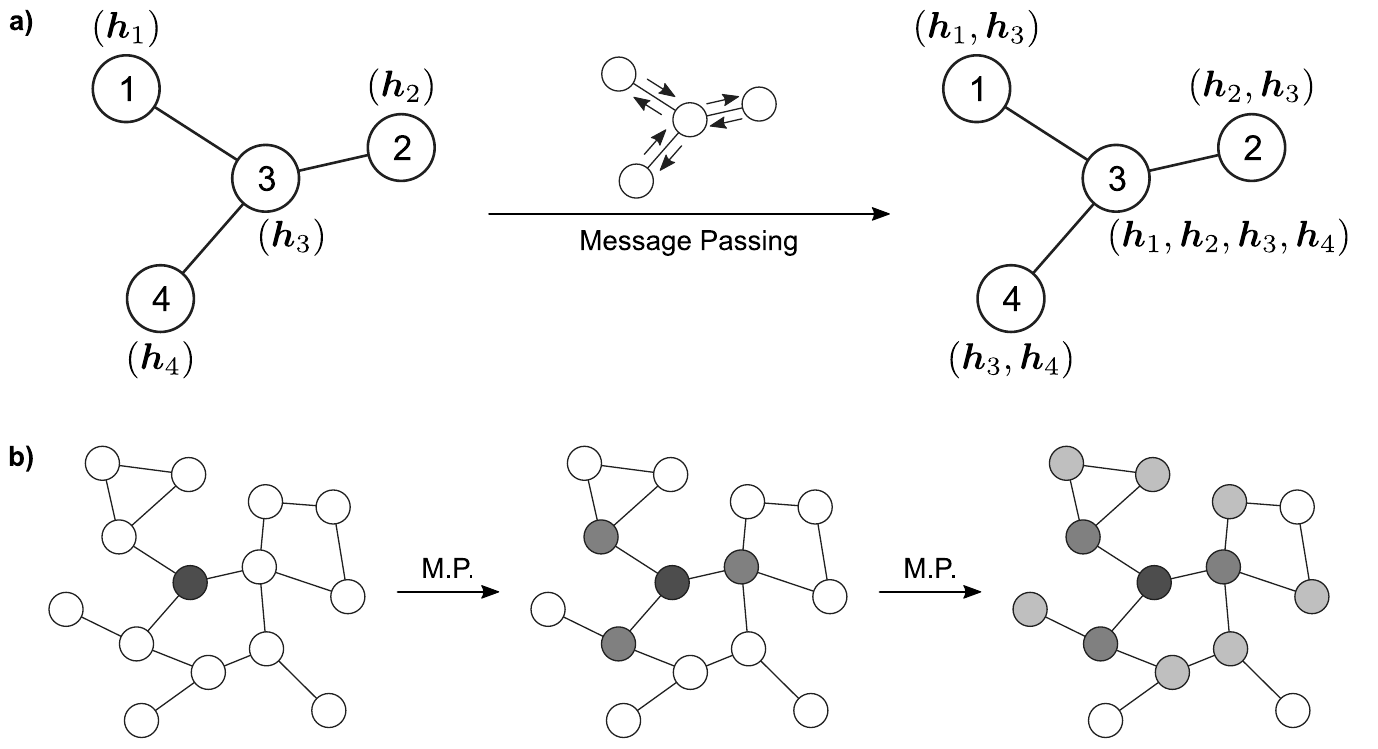}
    \caption{
        \textbf{a)} A schematic illustration of the message passing procedure.
        The $i$-th node has its original node information, $\bm h_i \ (i=1,\dots,4)$, at the beginning (left).
        The message passing procedure propagates node information between each pair of connected nodes (center).
        As a result, each node has its own information and neighbor information after the message passing (right).
        \textbf{b)} The scope of the information propagation expands along the message passing process.
        The black node in the middle of a graph has only its original node information at the beginning (left).
        This node obtains information in broader subgraphs through message passing, \IE dark gray nodes after one message passing step (center) and light gray nodes after two message passing steps (right).
        \textit{M.P.}: message passing.
    }
    \FigLab{intro}
\end{figure*}

In contrast to those classical graph machine learning methods using hand-crafted features, recent years have witnessed a surge in graph representation learning \citep[GRL;][]{hamilton2017representation}.
A GRL model learns a mapping from a node or a graph to a vector representation.
The mapping is trained so that the geometric relationships among embedded representations reflect the similarity of structural information in graphs, \IE nodes with similar local structures have similar representations \citep{belkin2001laplacian,ahmed2013distributed}.
The representation provided by the mapping can then be used as an input feature for task-specific models, such as classifiers or regressors.
Graph features \textit{learned} by GRL are more flexible than the \textit{hand-crafted} features used in classical graph machine learning methods.
However, the early GRL techniques learned a unique vector for each node without sharing parameters among nodes, leading to high computational costs and the risk of overfitting.
Furthermore, since these techniques learn a specific representation for each node, learned models cannot be applied for prediction on novel graphs or nodes that do not appear in the training phase \citep[Section~3.4]{hamilton2020graph}.

More recently, graph neural networks (GNNs) have rapidly emerged as a new framework for GRL (we refer readers to \citealt{zhang2018deep} and \citealt{wu2021comprehensive} for review papers; see \SecRef{survey} for related works).
Unlike non-GNN GRL techniques which learn node-specific representations, GNNs learn how to compute the node representation from the structural information around a node.
Hence, GNNs do not suffer from the problem that the computation cost increases linearly to the number of nodes.
Furthermore, the learned models generalize to graphs or nodes which are unknown while training.
The foundation of GNNs is the message passing procedure that propagates the information in a node to its neighbor nodes, each of which is directly connected to the source node by an edge (\FigRef{intro}a; see \SecRef{methods-gnn} for detail).
Since this procedure proceeds one step per layer, the range of the information propagation among nodes is small in the lower layers, and it expands toward the higher layers---\IE the node representations in the higher layers collect information from broader subgraphs (\FigRef{intro}b).

However, there is a problem in GNNs that the model performance degrades as the number of layers increases.
This is because deep GNN models lose the nodes' local information, which would be essential for good model performances, through many message passing steps.
This phenomenon is known as \textit{oversmoothing} \citep{li2018deeper}.
Many real-world graphs have fractal characteristics \citep{kim2007fractality}.
Therefore, a GRL model needs to be capable of capturing both local structural information and global structural information.
Capturing global structural information requires a GNN model to be deep (\IE having many message passing steps), but the oversmoothing problem prevents GNN models from getting deep.

In this study, we focus on improving learned graph representations for graph-level prediction tasks, such as molecular property classification.
Specifically, we seek a technique to learn more discriminative graph representation by using multiple representations in different localities.
Previous studies typically computed the graph representation by a graph pooling layer that collects node representations after the last message passing layer.
Therefore, deeper models cannot utilize nodes' local information in computing the graph representation because local information is lost through many message passing steps due to oversmoothing.
There are many prior works tackling the oversmoothing problem (see \SecRef{survey-deep}).
On the other hand, our approach---using information with multiple levels of localities to compute graph representations---does not aim to directly solve the oversmoothing problem itself, but we focus on improving the discriminability of learned representations.
To this end, we propose multi-level attention pooling (MLAP) architecture.
In short, the MLAP architecture introduces an attention pooling layer \citep{li2016gated} for each message passing step to compute layer-wise graph representations.
Then, it aggregates them to compute the final graph representation, inspired by the jumping knowledge network \citep{xu2018representation}.
Doing so, the MLAP architecture can focus on different nodes (or different subgraphs) in each layer with different levels of information localities, which leads to better modeling of both local structural information and global structural information.
In other words, introducing layer-wise attention pooling \textit{prior to} aggregating layer-wise representation would improve the graph-level classification performance.
Our experiments show performance improvements in GNN models with the MLAP architecture.
In addition, analyses on the layer-wise graph representations suggest that MLAP has the potential to learn graph representations with improved class discriminability by aggregating information with multiple levels of localities.

Our contributions in this work are as follows.
\begin{itemize}
    \item We propose the MLAP architecture for GNNs, which uses an attention-based global graph pooling \citep{li2016gated} for each message passing layer and the aggregation mechanism of layer-wise representations \citep{xu2018representation} in combination.
    \item We experimentally show that GNN models with MLAP architecture demonstrate better graph classification performance with multiple datasets.
    \item We also show the potential benefit of aggregating information in different levels of localities by analyzing layer-wise representations of MLAP.
\end{itemize}

The rest of this paper is organized as follows:
\SecRef{survey} summarizes related works,
\SecRef{methods} introduces the proposed MLAP framework,
\SecRef{experiments} describes the experimental setups,
\SecRef{results} demonstrates the results,
\SecRef{discussion} discusses the results,
and \SecRef{conclusion} concludes the present study.

\section{Related Works\SecLab{survey}}

\citet{gori2005new} and \citet{scarselli2009graph} first introduced the idea of GNNs, and \citet{bruna2014spectral} and \citet{defferrard2016convolutional} elaborated the formulation in the graph Fourier domain using spectral filtering.
Based on these earlier works, \citet{kipf2017semi} proposed the graph convolution network (GCN), which made a foundation of today's various GNN models \citep{duvenaud2015convolutional,hamilton2017inductive,niepart2016learning,velickovic2018graph,xu2019how}.
\citet{gilmer2017neural} summarized these methods as a framework named neural message passing, which computes node representations iteratively by collecting neighbor nodes' representation using differentiable functions.

In this study, we focus on methods to compute the graph representation from node-wise representations in GNN models.
We first summarize the studies on graph pooling methods and then review the recent trends in \textit{deep} GNN studies.
Finally, we summarize prior works that aggregate layer-wise representation to compute the final node/graph representation and elaborate the idea behind our proposed method.


\subsection{Graph Pooling Methods\SecLab{survey-pooling}}

Techniques to learn \textit{graph} representations are usually built upon those to learn \textit{node} representations.
A graph-level model first computes the representation for each node in a graph and then collects the node-wise representations into a single graph representation vector.
This collection procedure is called a pooling operation.
Although there are various pooling methods, they fall into two categories: the \textit{global} pooling approach and the \textit{hierarchical} pooling approach.

The \textit{global} pooling approach collects all of the node representations in a single computation.
The simplest example of the global pooling method is sum pooling, which merely computes the sum of all node representations.
\citet{duvenaud2015convolutional} introduced sum pooling to learn embedded representations of molecules from a graph where each node represents an atom.
Likewise, one can compute an average or take the maximum elements as a pooling method.
\citet{li2016gated} introduced attention pooling, which computes a weighted sum of node representations based on a softmax attention mechanism \citep{bahdanau2015neural}.
\citet{vinyals2016order} proposed set2set by extending the sequence to sequence (seq2seq) approach for a set without ordering.
\citet{zhang2018end} introduced the SortPooling, which sorts the node representations regarding their topological features and applies one-dimensional convolution.
These global pooling methods are simple and computationally lightweight, but they cannot use the structural information of graphs in the pooling operation.

In contrast, \textit{hierarchical} pooling methods segment the entire graph into a set of subgraphs hierarchically and compute the representations of subgraphs iteratively.
\citet{bruna2014spectral} introduced the idea of hierarchical pooling, or graph coarsening, based on hierarchical agglomerative clustering.
Although some early works like \citet{defferrard2016convolutional} also applied similar approaches, such clustering-based hierarchical pooling requires the clustering algorithm to be deterministic---that is, the hierarchy of subgraphs is fixed throughout the training.
To overcome this limitation, \citet{ying2018hierarchical} proposed DiffPool, which learns the subgraph hierarchy itself along with the message passing functions.
They proposed to use a neural network to estimate which subgraph a node should belong to in the next layer.
\citet{gao2019graph} extended U-Net \citep{ronneberger2015unet} for graph structure to propose graph U-Nets.
Original U-Net introduced down-sampling and up-sampling procedures for semantic image segmentation tasks.
Based on the U-Net, graph U-Nets are composed of a gPool network to shrink the graph size hierarchically and a gUnpool network to restore the original graph structure.
Also, \citet{lee2019self} employed a self-attention mechanism to define a hierarchy of subgraph structures.
Hierarchical pooling can adapt to multiple localities of graph substructures during step-wise shrinkage of graphs.
However, they are often computationally heavy because, as discussed in Cangea et al. (2018), they have to learn the dense \textit{assignment matrix} for each layer, relating a node in a layer to a node in the shrunk graph in the next layer.
Thus, they require longer computational time and consume larger memory.

\subsection{Oversmoothing in Deep Graph Neural Networks\SecLab{survey-deep}}

\citet{kipf2017semi} first reported that deep GNN models with many message passing layers performed worse than shallower models.
\citet{li2018deeper} investigated this phenomenon and found that deep GNN models converged to an equilibrium point wherein connected nodes have similar representations.
Since the nodes with similar representations are indistinguishable from each other, such convergence degrades the performance in node-level prediction tasks.
This problem is called \textit{oversmoothing}.
In graph-level prediction tasks, oversmoothing occurs independently for each graph.
Oversmoothing per graph damages GNN models' expressivity and results in performance degradation \citep{oono2020graph}.

Studies tackling the oversmoothing problem mainly fall into three categories: modifying the message passing formulation, adding residual connections, or normalization.
Anyhow, the objective of those studies is to retain discriminative representations even after many steps of message passing.

Studies modifying the message passing formulation aim to propose techniques to retain high-frequency components in graph signals during message passing steps, whereas message passing among nodes generally acts as a low-pass filter for the signals.
\citet{min2020scattering} proposed scattering GCN, which adds a circuit for band-pass filtering of node representations.
DropEdge \citep{rong2020dropedge} randomly removes some edges from the input graph, alleviating the low-pass filtering effect of the graph convolution.
Also, although not explicitly stated, the graph attention network \citep[GAT;][]{velickovic2018graph} is known to mitigate the oversmoothing problem because it can focus on specific nodes during message passing.

Adding residual connections is a more straightforward way to retain node-local representation up to deeper layers.
Residual connections, or ResNet architecture, were first introduced to convolutional neural networks for computer vision tasks, achieving a state-of-the-art performance \citep{he2016deep}.
\citet{kipf2017semi} applied the residual connections in the graph convolutional network and reported that residual connections mitigated the performance degradation in deeper models.
Later, \citet{li2019deepgcns}, \citet{zhang2019gresnet}, and \citet{chen2020simple} applied similar residual architectures on GNNs and showed performance improvement.

Normalization in deep learning gained attention by the success of early works such as BatchNorm \citep{ioffe2015batch} and LayerNorm \citep{ba2016layer}.
Although these general normalization techniques are also applicable and effective in GNNs, there are graph-specific normalization methods recently proposed.
PairNorm \citep{zhao2020pairnorm}, NodeNorm \citep{zhou2020understanding}, GraphNorm \citep{cai2020graphnorm}, and differentiable group normalization \citep[DGN;][]{zhou2020towards} are representative examples of graph-specific normalization methods.

These studies succeeded in overcoming the oversmoothing problem and making deep GNN models retain discriminative representations.
On the other hand, directly using local representations in computing the final graph representation would lead to more performance improvement.

\subsection{Aggregating Layer-Wise Representations in GNN}

The studies summarized in the previous subsection directly tackle the oversmoothing problem.
That is, they sought techniques to retain discriminative representations even after multiple steps of message passing.
Instead, we search for a technique to learn more discriminative representation by aggregating multiple representations in different localities.

Jumping knowledge (JK) network \citep{xu2018representation} proposed to compute the final node representation by aggregating intermediate layer-wise node representations.
Doing so, JK can adapt the locality of the subgraph from which a node gathers information.
After JK was proposed, many studies adopted JK-like aggregation of layer-wise representation to improve the learned representation.
\citet{wang2019kgat} adopted JK in recommendation tasks on knowledge graphs.
\citet{cangea2018towards} adopted a JK-like aggregation of layer-wise pooled representation upon gPool \citep{gao2019graph} network to learn graph-level tasks.
A similar combination of hierarchical graph pooling and JK-like aggregation was also proposed by \citet{ranjan2020}.
\citet{dehmamy2019understanding} proposed aggregating layer-wise representation from a modified GCN architecture and showed performance improvement.

Our proposed MLAP technique is motivated by the same idea of these studies that GNNs should be capable of aggregating information in multiple levels of localities.
Here, we utilize an intuition on graph-level prediction tasks: a model should focus on different nodes as the message passing proceeds through layers and the locality of information extends.
That is, the importance of a node in global graph pooling would differ depending on the locality of the information.
Therefore, in this study, we propose a method that uses an attention-based global pooling in each layer and aggregates those layer-wise graph representations to compute the final graph representation.

\section{Methods\SecLab{methods}}

We propose the MLAP architecture, which aggregates graph representation from multiple levels of localities.
In this section, we first summarize the fundamentals of GNNs, particularly the message passing procedure, and then introduce the MLAP architecture.

\begin{figure*}[t]
    \centering
    \includegraphics{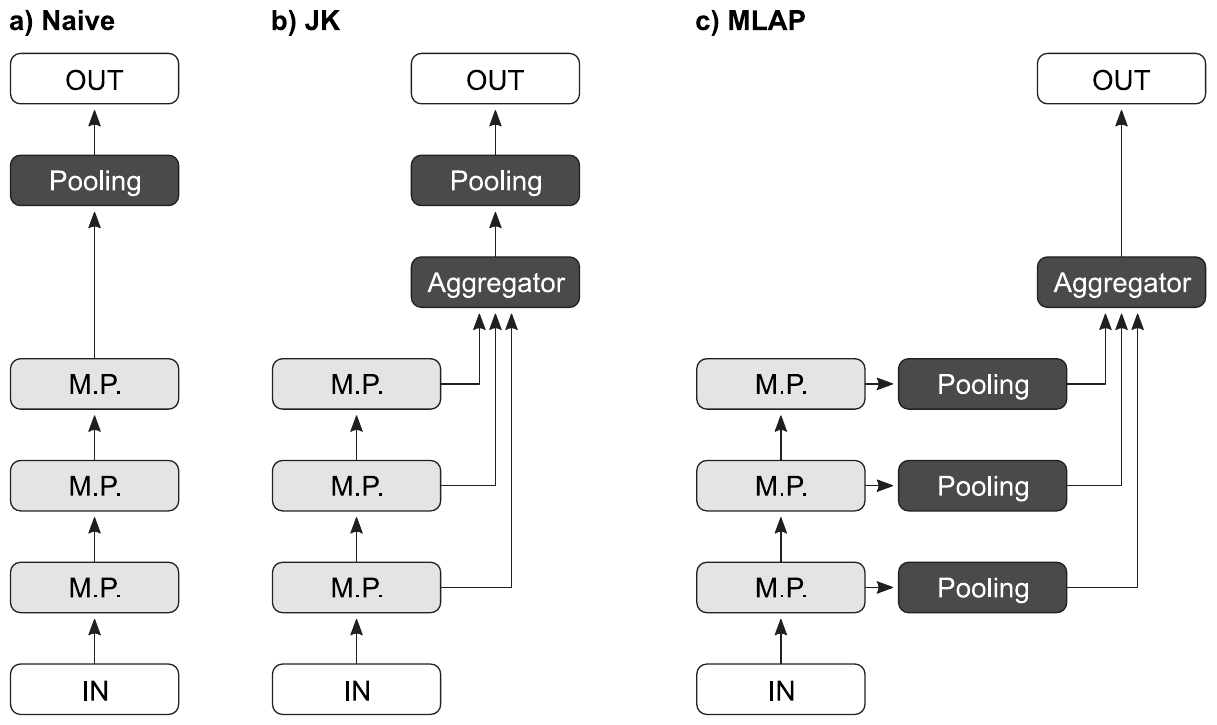}
    \caption{
        \textbf{a)} A naive GNN architecture.
        A pooling layer computes the graph representation from the node representations after the last message passing.
        \textbf{b)} The jumping knowledge (JK) network architecture.
        The aggregator collects the layer-wise \textit{node} representation, and then a pooling layer computes the graph representation from the aggregated node representation.
        \textbf{c)} Proposed multi-level attentional pooling (MLAP) architecture.
        There is a dedicated pooling layer for each message passing layer to compute layer-wise \textit{graph} representation.
        The aggregator computes the final graph representation from the layer-wise graph representations.
        \textit{M.P.}: message passing.}
    \FigLab{arch}
\end{figure*}


\subsection{Preliminaries: Graph Neural Networks\SecLab{methods-gnn}}
Let $\MlapGraph = (\MlapNodes, \MlapEdges)$ be a graph, where $\MlapNodes$ is a set of nodes and $\MlapEdges$ is a set of edges.
$\MlapNode \in \MlapNodes$ denotes a node and $\MlapEdge{\MlapNodeSrc}{\MlapNodeDst} \in \MlapEdges$ denotes a directed edge from a source node $\MlapNodeSrc$ to a destination node $\MlapNodeDst$.
A graph may have node features or edge features, or both of them.
If a graph has node features, each node $\MlapNode$ has a node feature vector $\MlapNodeFeat{\MlapNode}$.
Similarly, if a graph has edge features, each edge $\MlapEdge{\MlapNodeSrc}{\MlapNodeDst}$ has an edge feature vector $\MlapEdgeFeat{\MlapNodeSrc}{\MlapNodeDst}$.

There are three types of tasks commonly studied for GNNs: graph-level prediction, node-level prediction, and edge-level prediction.
In this study, we focus on the graph-level prediction tasks, that is,
given a set of graph $\MlapGraphs = \left\{ \MlapGraph_1, \dots, \MlapGraph_{|\MlapGraphs|} \right\}$ and their labels $\MlapLabels = \left\{ \MlapLabel_1, \dots, \MlapLabel_{|\MlapGraphs|} \right\}$,
we want to learn a graph representation vector $\MlapGraphEmb{\MlapGraph}$ used for predicting the graph label $\hat\MlapLabel_\MlapGraph = g\left( \MlapGraphEmb{\MlapGraph} \right)$, where $g$ is a predictor function.

Suppose we have a GNN with $\MlapNumLayer$ layers.
Each layer in a GNN propagates the node representation $\MlapNodeEmbBare{\MlapNode}$ along the edges (\textit{message passing}).
Let $\MlapNodeEmb{\MlapNode}{\MlapLayer} \in \Real^{\MlapEmbDim}$ be the representation of $\MlapNode$ after the message passing by the $\MlapLayer$-th layer, where $\MlapEmbDim$ is the dimension of the vector representations.
In general, the propagation by the $l$-th layer first computes the \textit{message} $\MlapMsg{\MlapNode}{\MlapLayer}$ for each node $\MlapNode$ from its neighbor nodes $\MlapNeighbor{\MlapNode}$, as in
\begin{equation}
    \MlapMsg{\MlapNode}{\MlapLayer} =  \MlapColFn{\MlapLayer} \left( \left\{ \MlapMsgFn{\MlapLayer} \left( \MlapNodeEmb{\MlapNodePrime}{\MlapLayer-1}, \MlapEdgeFeat{\MlapNodePrime}{\MlapNode} \right) \ \middle| \ \MlapNodePrime \in \MlapNeighbor{\MlapNode} \right\} \right), \EqLab{gnn-msg}
\end{equation}
where $\MlapMsgFn{\MlapLayer}$ is a message function to compute the message for each neighbor node from the neighbor representation and the feature of the connecting edge, and $\MlapColFn{\MlapLayer}$ is a function to collect the neighbor node-wise messages.
Then, the layer \textit{updates} the node representation $\MlapNodeEmb{\MlapNode}{\MlapLayer}$ as
\begin{equation}
    \MlapNodeEmb{\MlapNode}{\MlapLayer} = \MlapUpdFn{\MlapLayer} \left( \MlapMsg{\MlapNode}{\MlapLayer}, \MlapNodeEmb{\MlapNode}{\MlapLayer-1} \right), \EqLab{gnn-upd}
\end{equation}
where $\MlapUpdFn{\MlapLayer}$ is an update function.

After $\MlapNumLayer$ steps of message passing, a graph pooling layer computes a graph representation vector $\MlapGraphEmb{\MlapGraph}$ from the final node representations $\MlapNodeEmb{\MlapNode}{\MlapNumLayer}$ for each $\MlapNode \in \MlapNodes$, as in
\begin{equation}
    \MlapGraphEmb{\MlapGraph} = \MlapPool \left( \left\{ \MlapNodeEmb{\MlapNode}{\MlapNumLayer} \ \middle| \ \MlapNode \in \MlapNodes \right\} \right).
    \EqLab{single-pool}
\end{equation}


\subsection{Multi-Level Attentional Pooling}

Graph-level prediction tasks require the models to utilize both local information in nodes and global information as the entire graphs for good performances.
However, typical GNN implementations first execute the message passing among nodes for a certain number of steps $\MlapNumLayer$ \textit{and then} pool the node representations into a graph representation, as shown in \EqRef{single-pool} (\FigRef{arch}a).
This formulation damages GNN models' expressivity because it can only use the information in a fixed locality to compute the graph representation.

To fix this problem, we introduce a novel GNN architecture named multi-level attentional pooling (MLAP; \FigRef{arch}c).
In the MLAP architecture, each message passing layer has a dedicated pooling layer to compute layer-wise graph representations, as in
\begin{equation}
    \MlapLayerEmb{\MlapGraph}{\MlapLayer} = \MlapLayerPool{\MlapLayer} \left( \left\{ \MlapNodeEmb{\MlapNode}{\MlapLayer} \ \middle| \ \MlapNode \in \MlapNodes \right\} \right) \quad \forall \ \MlapLayer \in \{1, \dots, \MlapNumLayer\}.
\end{equation}
Here, we used the attention pooling \citep{li2016gated} as the pooling layer.
Thus,
\begin{align}
    \MlapLayerEmb{\MlapGraph}{\MlapLayer} &= \sum_{\MlapNode \in \MlapNodes} \mathrm{softmax} \left( \MlapLayerGateFn{\MlapLayer} (\MlapNodeEmb{\MlapNode}{\MlapLayer}) \right) \MlapNodeEmb{\MlapNode}{\MlapLayer} \EqLab{mlap-hl} \\
                                          &= \sum_{\MlapNode \in \MlapNodes}
                                                \frac{
                                                    \exp \left( \MlapLayerGateFn{\MlapLayer} (\MlapNodeEmb{\MlapNode}{\MlapLayer}) \right)
                                                }{
                                                    \sum_{\MlapNodePrime \in \MlapNodes} \exp \left( \MlapLayerGateFn{\MlapLayer} (\MlapNodeEmb{\MlapNodePrime}{\MlapLayer}) \right)
                                                }
                                                \MlapNodeEmb{\MlapNode}{\MlapLayer},
\end{align}
where $\MlapLayerGateFn{\MlapLayer}$ is a function used to compute the attention score, for which we used a two-layer neural network.

Then, an aggregation function computes the final graph representation by unifying the layer-wise representations as follows:
\begin{equation}
    \MlapGraphEmb{\MlapGraph} = \MlapAggFn \left( \left\{ \MlapLayerEmb{\MlapGraph}{\MlapLayer} \ \middle| \ \MlapLayer \in \{1, \dots, \MlapNumLayer\} \right\} \right),
\end{equation}
where $\MlapAggFn$ is an aggregation function.
One can use an arbitrary function for $\MlapAggFn$.
In the present study, we tested two types of aggregation functions: \textit{Sum} and \textit{Weighted}.

\subsubsection{Sum}
One of the simplest ways to aggregate the layer-wise graph representations is to take the sum of them, as in
\begin{equation}
    \MlapGraphEmb{\MlapGraph} = \sum_{\MlapLayer=1}^\MlapNumLayer \MlapLayerEmb{\MlapGraph}{\MlapLayer}.
    \EqLab{mlap-sum}
\end{equation}
This formulation expresses an assumption that the representation in each layer is equally important in computing the final graph representation.

\subsubsection{Weighted}
Each layer-wise representation might have different importance depending on the layer index.
If this is the case, taking a weighted sum would be adequate to learn such importance of layers, as in
\begin{equation}
    \MlapGraphEmb{\MlapGraph} = \sum_{\MlapLayer=1}^\MlapNumLayer \MlapWeight{\MlapLayer} \MlapLayerEmb{\MlapGraph}{\MlapLayer},
    \EqLab{mlap-weighted}
\end{equation}
where $\left\{ \MlapWeight{\MlapLayer} \ \middle| \ \MlapLayer \in \{1, \dots, \MlapNumLayer\} \right\}$ is a trainable weight vector.

\section{Experiments\SecLab{experiments}}

Our experimental evaluation aims to answer these research questions:
\begin{enumerate}[RQ1]
    \item Does the MLAP architecture improve the GNN performances in graph classification tasks?
    \item Does aggregating representations from multiple layers have a benefit in learning discriminative graph representation?
\end{enumerate}
To this end, we conducted experiments using four graph classification datasets: a synthetic dataset and three real-world datasets.


\subsection{Synthetic Dataset\SecLab{experiments-synthetic}}

\begin{figure*}[t]
    \centering
    \includegraphics{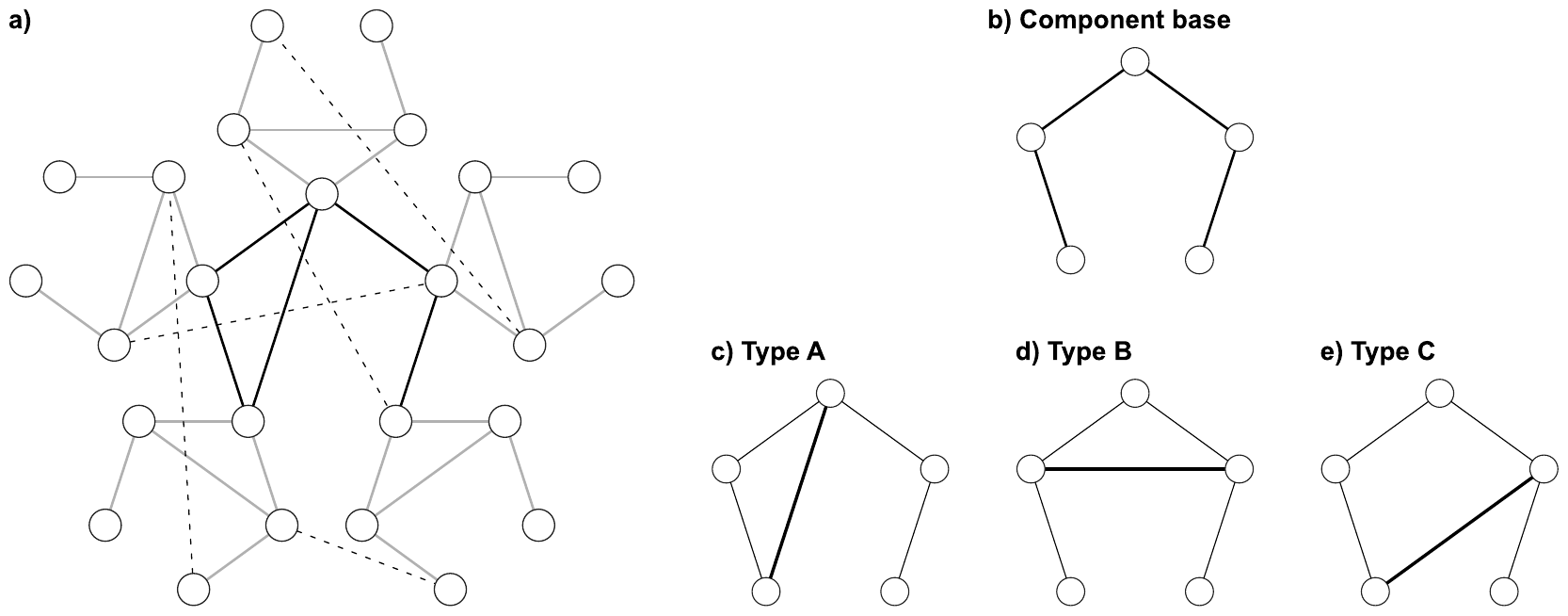}
    \caption{
        \textbf{a)} A graph in the synthetic dataset.
        It consists of the \textit{center} component (black edges), five \textit{peripheral} components (gray edges), and five additional random edges (dotted edges).
        The class of this graph is determined by the combination of the types of the center component (type A) and the peripheral components (type B).
        \textbf{b)} The basic structure of a component.
        \textbf{c--e)} Three types of components.
    }
    \FigLab{synthetic}
\end{figure*}

We created a synthetic dataset to show the effectiveness of MLAP using multi-level representation in a graph-level classification task.
We designed the dataset so that its graph features are represented in both local and global graph structures.

A graph in the dataset consists of six 5-node components: one \textit{center} component surrounded by five \textit{peripheral} components, each of which shares a node with the center component (\FigRef{synthetic}a).
The basic structure of a component is five sequentially connected nodes (\FigRef{synthetic}b) with an extra edge.
Depending on how the extra edge is appended, there are three types of components (\FigRef{synthetic}c--e).
The class of a graph is determined by the combination of the type of the center component and the type of the peripheral components.
Note that the five peripheral components share the same type.
Therefore, there are $3\times3=9$ classes of graphs.
By this design, accurately classifying the graphs in this dataset requires a model to learn both the local substructures in a graph and the global structure as an entire graph (\IE the combination of the types of local substructures).
Neither nodes nor edges in the graphs have features.

We generated 1,000 unique graphs for each class by randomly appending five edges between arbitrarily selected pairs of nodes.
Hence, there were 9,000 instances in the dataset in total, and we applied a random 8:1:1 split to provide training, validation, and test sets.
Model performance was evaluated by the error rate ($1-\mathrm{Accuracy}$).


\subsection{Real-World Benchmark Datasets\SecLab{experiments-datasets}}
We used two datasets from the open graph benchmark \citep[OGB;][]{hu2020open} and the MCF-7 dataset from the TU graph dataset collection \citep{morris2020tudataset}.

\subsubsection{ogbg-molhiv}
ogbg-molhiv is a dataset for a molecular property prediction task, originally introduced in \citet{wu2018moleculenet}.
Each graph in the dataset represents a molecule.
Each node in a graph represents an atom and has a 9-dimensional discrete-valued feature containing the atomic number and other atomic properties.
Each edge represents a chemical bond between two atoms and has a 3-dimensional discrete-valued feature containing the bond type and other properties.
This dataset has a relatively small sample size (41,127 graphs in total), with 25.5 nodes and 27.5 edges per graph on average.
The task is a binary classification to identify whether a molecule inhibits the human immunodeficiency virus (HIV) from replication.
Model performance is evaluated by the area under the curve value of the radar operator characteristics curve (ROC-AUC).
We followed the standard dataset splitting procedure provided by the OGB.

\subsubsection{ogbg-ppa}
The ogbg-ppa dataset contains a set of subgraphs extracted from protein-protein association networks of species in 37 taxonomic groups, originally introduced in \citet{szklarczyk2018string}.
Each node in a graph represents a protein without node features.
Each edge represents an association between two proteins and has a 7-dimensional real-valued feature describing the biological meanings of the association.
This dataset has a medium sample size (158,100 graphs in total), with 243.4 nodes and 2266.1 edges per graph on average.
The task is a classification to identify from which taxonomic group among 37 classes an association graph comes.
The performance of a model is evaluated by the overall classification accuracy.
We followed the standard dataset splitting procedure provided by the OGB.

\subsubsection{MCF-7}
MCF-7 is a chemical molecule dataset originally extracted from PubChem\footnote{https://pubchem.ncbi.nlm.nih.gov}.
Each graph in the dataset represents a molecule, and the task is a binary classification of whether a molecule inhibits the growth of a human breast tumor cell line.
Each node in a graph represents an atom and has a 1-dimensional discrete-valued feature describing the atomic number.
Each edge represents a chemical bond between two atoms and has a 1-dimensional discrete-valued feature describing the bond type.
This dataset has a relatively small sample size (27,770 graphs in total), with 26.4 nodes and 28.5 edges per graph on average.
Model performance is evaluated by ROC-AUC.
Since the TU dataset does not provide a standard data split, we applied a random 8:1:1 split into training, validation, and test sets.


\subsection{Model Configurations}
We used the graph isomorphism network \citep[GIN;][]{xu2019how} as the message passing layer\footnote{Note that the MLAP architecture is applicable to any GNN models independent of the type of message passing layers.} following the OGB's reference implementation shown in \citet{hu2020open}, \IE in Eqs.~\EqRefBare{gnn-msg} and \EqRefBare{gnn-upd},
\begin{align}
    \MlapMsg{\MlapNode}{\MlapLayer} &= \sum_{\MlapNodePrime \in \MlapNeighbor{\MlapNode}} \mathrm{ReLU}\left( \MlapNodeEmb{\MlapNodePrime}{\MlapLayer - 1} + \MlapEdgeFn{\MlapLayer}(\MlapEdgeFeat{\MlapNodePrime}{\MlapNode}) \right), \EqLab{gin-msg} \\
    \MlapNodeEmb{\MlapNode}{\MlapLayer} &= \MlapGinNnFn{\MlapLayer} \left( (1 + \MlapGinEps{\MlapLayer})\cdot \MlapNodeEmb{\MlapNode}{\MlapLayer-1} + \MlapMsg{\MlapNode}{\MlapLayer} \right),
\end{align}
where $\MlapEdgeFn{\MlapLayer}$ is a trainable function to encode edge features into a vector, $\MlapGinNnFn{\MlapLayer}$ is a two-layer neural network for transforming node representations, and $\MlapGinEps{\MlapLayer}$ is a trainable scalar weight modifier.

We varied the number of GIN layers $\MlapNumLayer$ from 1 to 10 to investigate the effect of depth in model performance.
We fixed the node representation dimension $\MlapEmbDim$ to 200 and added a dropout layer for each GIN layer with a dropout ratio of 0.5.
\textcolor{red}{In addition, each message passing operation is followed by a GraphNorm operation \citep{cai2020graphnorm} before dropout under the GraphNorm (+) configuration.}
We optimized the model using the Adam optimizer \citep{kingma2015adam}.

There are dataset-specific settings detailed below.

\subsubsection{Synthetic Dataset}
Since the graphs in the synthetic dataset do not have the node features nor edge features, we set $\MlapNodeFeat{\MlapNode} = 0$ and $\MlapEdgeFeat{\MlapNodeSrc}{\MlapNodeDst} = 0$.
Each GIN layer had an edge feature encoder that returned a constant $\MlapEmbDim$-dimensional vector.

Besides GNN, each model learned an embedded class representation matrix $\MlapPPAClassEmbMat \in \Real^{9 \times \MlapEmbDim}$.
The probability with which a graph belongs to the class $c$ was computed by a softmax function:
\begin{equation}
    P (c|\MlapGraph) = \mathrm{softmax} \left( \MlapPPAClassEmb{c} \cdot \MlapGraphEmb{\MlapGraph} \right)
                     = \frac{
                            \exp \left( \MlapPPAClassEmb{c} \cdot \MlapGraphEmb{\MlapGraph} + b_c \right)
                        }{
                            \sum_{c'=1}^{9} \exp \left( \MlapPPAClassEmb{c'} \cdot \MlapGraphEmb{\MlapGraph} + b_{c'} \right)
                        },
    \EqLab{classifier-multi}
\end{equation}
where $\MlapPPAClassEmb{c}$ is the $c$-th row vector of $\MlapPPAClassEmbMat$, and $b_c$ is the bias term for the class $c$.

The models were trained against a cross-entropy loss function for 65 epochs.
The initial learning rate was set to $10^{-3}$ and decayed by $\times 0.2$ for every 15 epochs.
The batch size was 50.

\subsubsection{ogbg-molhiv}
We used the OGB's atom encoder for computing the initial node representation $\MlapNodeEmb{\MlapNode}{0}$ from the 9-dimensional node feature.
We also used the OGB's bond encoder as $\MlapEdgeFn{\MlapLayer}$, which takes the 3-dimensional edge feature as its input.

After computing the graph representation $\MlapGraphEmb{\MlapGraph}$, a linear transformation layer followed by a sigmoid function computes the probability with which each graph belongs to the \textit{positive} class, as in
\begin{equation}
    P (\mathrm{positive}|\MlapGraph) = \sigma \left( \MlapMolhivOutWeight \cdot \MlapGraphEmb{\MlapGraph} + b \right),
    \EqLab{classifier-molhiv}
\end{equation}
where $\sigma$ is a sigmoid function, and $\MlapMolhivOutWeight$ is a trainable row vector with the same dimension $\MlapEmbDim$ as the graph representation vectors.
$b$ is the bias term.

The models were trained against a binary cross-entropy loss function for 50 epochs.
The initial learning rate was set to $10^{-4}$ and decayed by $\times 0.5$ for every 15 epochs.
The batch size was set to 20 to avoid overfitting.

\subsubsection{ogbg-ppa}
We set $\MlapNodeFeat{\MlapNode} = 0$ because this dataset does not have node features.
We used a two-layer neural network as $\MlapEdgeFn{\MlapLayer}$ to embed the edge feature.

The multi-class classification procedure and the hyperparameters for optimization are identical to those used for the synthetic dataset, except that the number of classes was 37 and that the models were trained for 50 epochs.

\subsubsection{MCF-7}
We trained a vector embedding for node features as $\MlapNodeEmb{\MlapNode}{0}$ and another vector embedding for $\MlapEdgeFn{\MlapLayer}$ in \EqRef{gin-msg}.

The binary classification procedure and the hyperparameters for optimization were identical to those used for ogbg-molhiv.


\subsection{Performance Evaluation (RQ1)}

\subsubsection{Baseline Models}

We compared the performance of GNN models with our MLAP framework (\FigRef{arch}c) to two baseline models.
One was a naive GNN model that simply stacked GIN layers, wherein the representation of a graph was computed by pooling the node representations after the last message passing (\FigRef{arch}a), as in
\begin{equation}
    \MlapGraphEmb{\MlapGraph} = \MlapPool \left( \left\{ \MlapNodeEmb{\MlapNode}{\MlapNumLayer} \ \middle| \ \MlapNode \in \MlapNodes \right\} \right).
\end{equation}
We used the same attention pooling as MLAP, that is,
\begin{equation}
    \MlapGraphEmb{\MlapGraph} = \sum_{\MlapNode \in \MlapNodes} \mathrm{softmax} \left( \MlapGateFn (\MlapNodeEmb{\MlapNode}{\MlapNumLayer}) \right) \MlapNodeEmb{\MlapNode}{\MlapNumLayer}.
\end{equation}

The other was the JK architecture \citep{xu2018representation}, which first computed the final node representations by aggregating layer-wise node representations, and the graph representation was computed by pooling the aggregated node representations \citep[\FigRef{arch}b;][]{xu2019how}, as in
\begin{equation}
    \MlapGraphEmb{\MlapGraph} = \MlapPool \left(\left\{ \MlapNodeEmb{\MlapNode}{\mathrm{JK}} \ \middle| \ \MlapNode \in \MlapNodes \right\}\right).
\end{equation}
Here, $\MlapNodeEmb{\MlapNode}{\mathrm{JK}}$ is the aggregated node representation computed from the layer-wise node representation, as in
\begin{equation}
    \MlapNodeEmb{\MlapNode}{\mathrm{JK}} = \MlapJK \left(\left\{ \MlapNodeEmb{\MlapNode}{\MlapLayer} \ \middle| \ \MlapLayer \in \{1, \dots, \MlapNumLayer\} \right\}\right),
    \EqLab{jk-hjk}
\end{equation}
where $\MlapJK$ is the JK's aggregation function, for which
we tested all three variants proposed in \citet{xu2018representation}---\textit{Concatenation}, \textit{MaxPool}, and \textit{LSTM-Attention}---and \textit{Sum} used in the OGB's reference implementation, defined as
\begin{equation}
    \MlapJK \left(\left\{ \MlapNodeEmb{\MlapNode}{\MlapLayer} \ \middle| \ \MlapLayer \in \{1, \dots, \MlapNumLayer\} \right\}\right) =
        \sum_{\MlapLayer=1}^{\MlapNumLayer} \MlapNodeEmb{\MlapNode}{\MlapLayer}.
    \EqLab{jk-sum}
\end{equation}
Finally, the graph representation was computed using the attention pooling, as in
\begin{equation}
    \MlapGraphEmb{\MlapGraph} = \sum_{\MlapNode \in \MlapNodes} \mathrm{softmax} \left( \MlapGateFn (\MlapNodeEmb{\MlapNode}{\mathrm{JK}}) \right) \MlapNodeEmb{\MlapNode}{\mathrm{JK}}.
    \EqLab{jk-pool}
\end{equation}

For each architecture, we trained models with varying depth (1--10).

\subsubsection{Statistical Analyses}

We trained models using 30 different random seeds, except that we used 10 seeds for ogbg-ppa because the dataset is bigger than others and takes a long time for training.
The performance of an architecture with a certain depth was evaluated by the mean and the standard error.

\textcolor{red}{We selected the best model configuration with regard to the depth, the type of aggregator, and GraphNorm (+) or ($-$) among each of the naive, JK, MLAP architecture.
Then}, we compared the performance of the best MLAP model to the best naive models and the best JK models using Mann-Whitney $U$-test.
Also, we computed the effect size.
Given the test statistic $z$ from the $U$-test, the effect size $r$ was computed as $r = z / \sqrt{N}$, where $N$ is the total number of samples.


\begin{figure*}[t]
    \centering
    \includegraphics{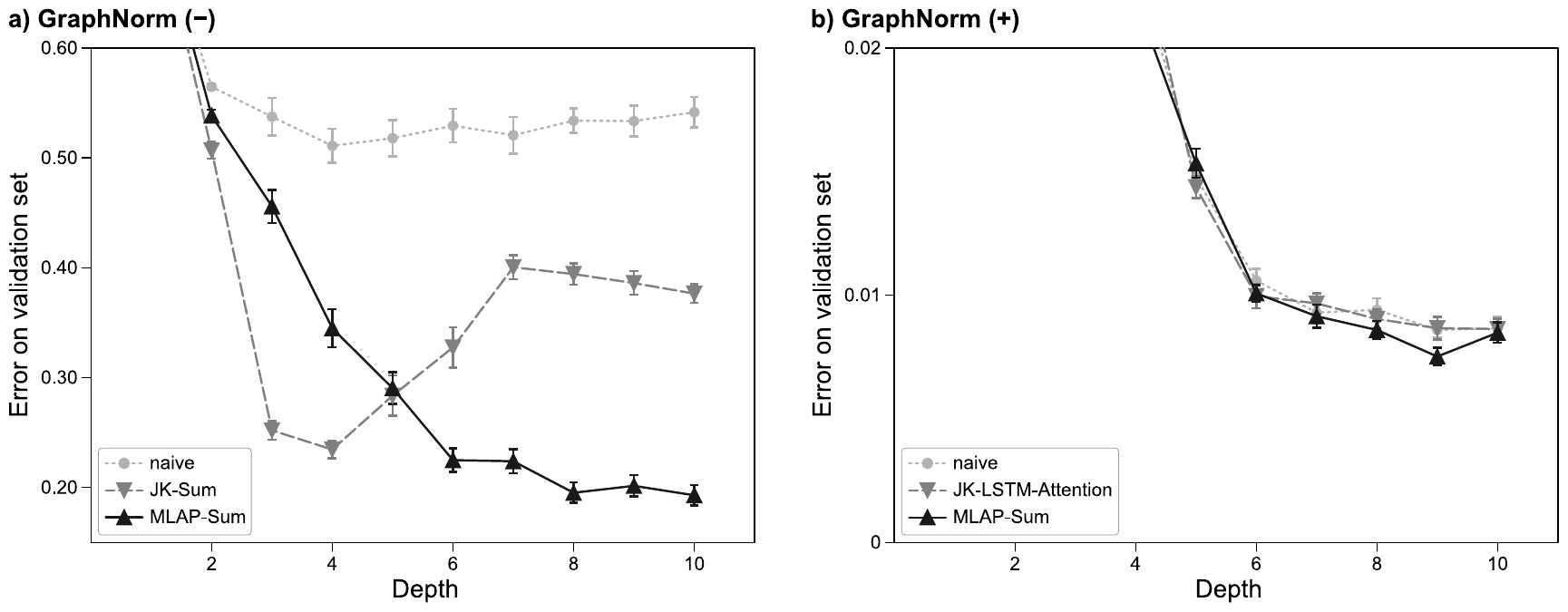}
    \caption{
        The \textcolor{red}{validation} performances for the synthetic dataset.
        Full results are in Appendix \textcolor{red}{\TblRef{valid-full-fractal}}.
    }
    \FigLab{results-fractal-depth-perf}
\end{figure*}

\begin{figure*}[t]
    \centering
    \includegraphics{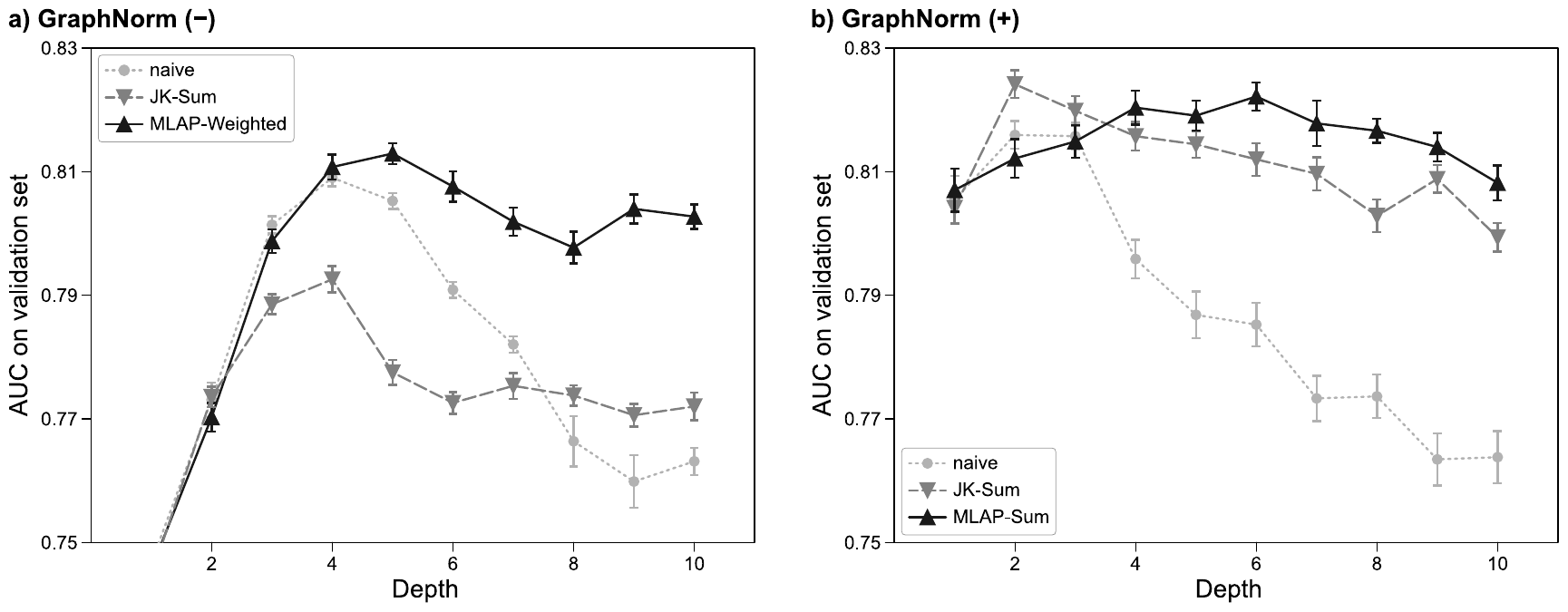}
    \caption{
        The \textcolor{red}{validation} performances for the ogbg-molhiv dataset.
        Full results are in Appendix \textcolor{red}{\TblRef{valid-full-molhiv}}.
    }
    \FigLab{results-molhiv-depth-perf}
\end{figure*}

\begin{figure*}[t]
    \centering
    \includegraphics{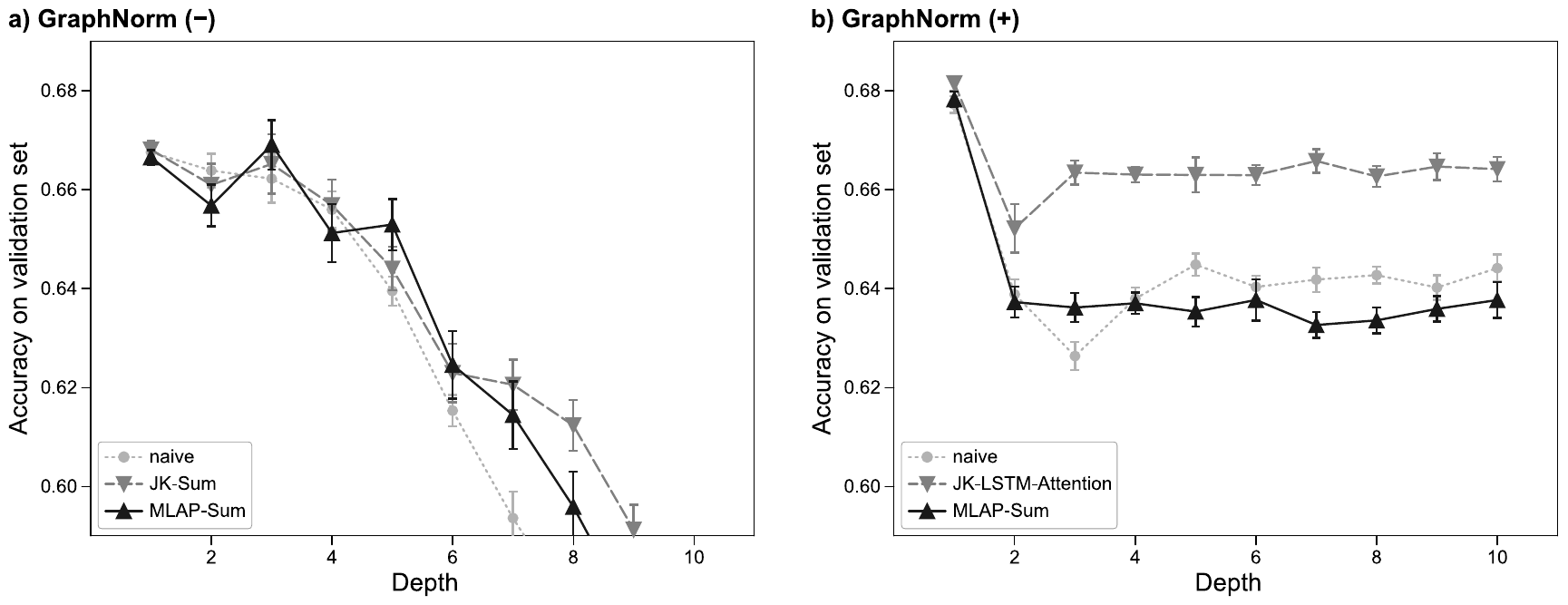}
    \caption{
        The \textcolor{red}{validation} performances for the ogbg-ppa dataset.
        Full results are in Appendix \textcolor{red}{\TblRef{valid-full-ppa}}.
    }
    \FigLab{results-ppa-depth-perf}
\end{figure*}

\begin{figure*}[t]
    \centering
    \includegraphics{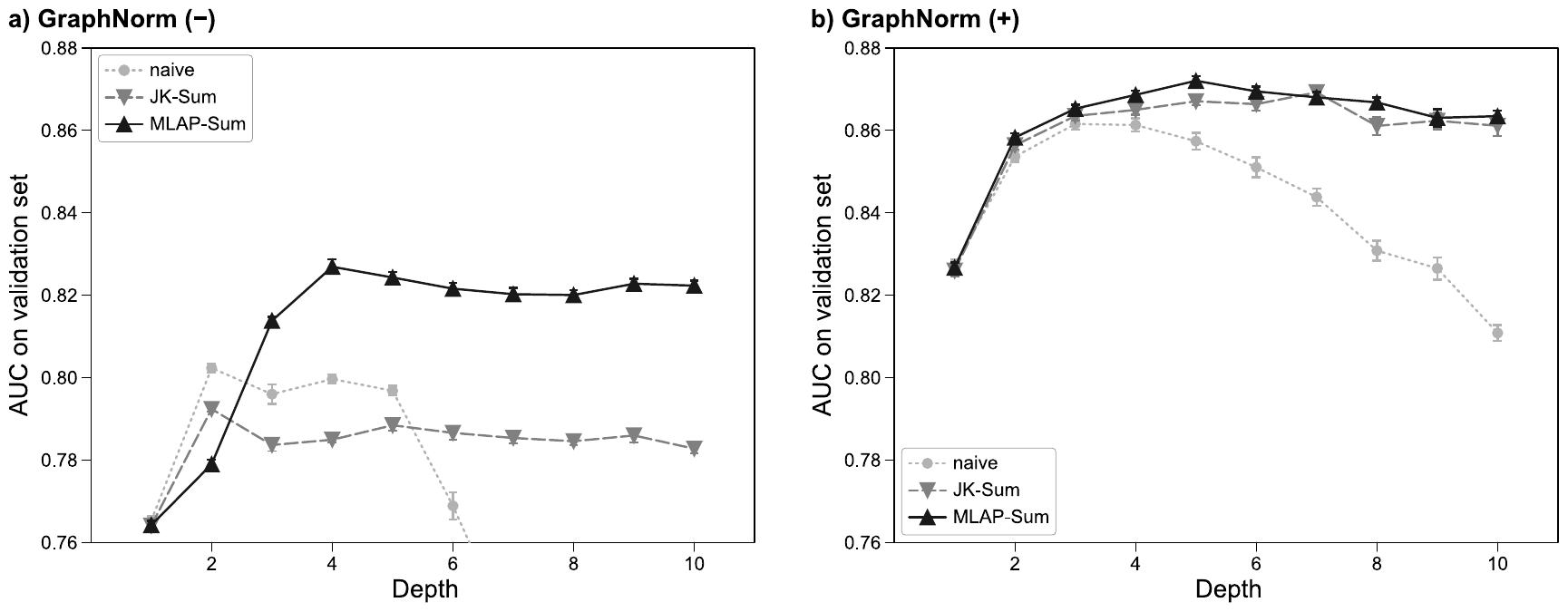}
    \caption{
        The \textcolor{red}{validation} performances for the MCF-7 dataset.
        Full results are in Appendix \textcolor{red}{\TblRef{valid-full-mcf}}.
    }
    \FigLab{results-mcf-depth-perf}
\end{figure*}


\begin{table*}[t]
    \small
    \centering
    \begin{tabular}{l||ccc|ccc|ccc|ccc}
                               & \multicolumn{3}{c|}{Synthetic}           & \multicolumn{3}{c|}{ogbg-molhiv}         & \multicolumn{3}{c|}{ogbg-ppa}            & \multicolumn{3}{c}{MCF-7}               \\
                               & Aggregator & ($L$) & GN                  & Aggregator & ($L$) & GN                  & Aggregator & ($L$) & GN                  & Aggregator & ($L$) & GN                 \\
        Base                   & \multicolumn{3}{c|}{Test perf.}          & \multicolumn{3}{c|}{Test perf.}          & \multicolumn{3}{c|}{Test perf.}          & \multicolumn{3}{c}{Test perf.}          \\ \hline\hline
        \multirow{2}{*}{naive} & -          & (9)   & (+)                 & -          & (2)   & (+)                 & -          & (1) & (+)                   & -          & (3)   & (+)                \\
                               & \multicolumn{3}{c|}{0.0175 $\pm$ 0.0007} & \multicolumn{3}{c|}{0.7567 $\pm$ 0.0034} & \multicolumn{3}{c|}{0.7184 $\pm$ 0.0011} & \multicolumn{3}{c}{0.8572 $\pm$ 0.0012} \\ \hline
        \multirow{2}{*}{JK}    & LSTM-Att.  & (10)  & (+)                 & Sum        & (2)   & (+)                 & LSTM-Att.  & (1) & (+)                   & Sum        & (7)   & (+)                \\
                               & \multicolumn{3}{c|}{0.0163 $\pm$ 0.0005} & \multicolumn{3}{c|}{0.7708 $\pm$ 0.0030} & \multicolumn{3}{c|}{0.7198 $\pm$ 0.0013} & \multicolumn{3}{c}{0.8572 $\pm$ 0.0012} \\ \hline
        \multirow{2}{*}{MLAP}  & Sum        & (9)   & (+)                 & Sum        & (6)   & (+)                 & Sum        & (1) & (+)                   & Sum        & (5)   & (+)                \\
                               & \multicolumn{3}{c|}{0.0150 $\pm$ 0.0006} & \multicolumn{3}{c|}{0.7651 $\pm$ 0.0027} & \multicolumn{3}{c|}{0.7183 $\pm$ 0.0012} & \multicolumn{3}{c}{0.8634 $\pm$ 0.0011}
    \end{tabular}
    \caption{
        The test performances (\textit{mean} $\pm$ \textit{standard error}) of the selected models.
        We chose the best combination of the aggregator and the model depth.
        GN: GraphNorm; LSTM-Att.: LSTM-Attention.
    }
    \TblLab{results-test}
\end{table*}

\begin{table*}[t]
    \small
    \centering
    \begin{tabular}{l||cc|cc|cc|cc}
                       & \multicolumn{2}{c|}{Synthetic} & \multicolumn{2}{c|}{ogbg-molhiv} & \multicolumn{2}{c|}{ogbg-ppa} & \multicolumn{2}{c}{MCF-7} \\
        Comparison     & $p$         & E.S.             & $p$    & E.S.                    & $p$   & E.S.                  & $p$    & E.S.             \\ \hline\hline
        MLAP vs. naive & *0.004      & 0.345            & *0.039 & 0.229                   & 0.367 & $-$0.085              & *$<10^{-3}$ & 0.418       \\
        MLAP vs. JK    & *0.039      & 0.226            & 0.120  & $-$0.153                & 0.192 & $-$0.203              & *$<10^{-3}$ & 0.407
    \end{tabular}
    \caption{
        The statistical analysis results. We compared the best performance among MLAP models to naive models and JK models.
        $p$: $p$-value of the Mann-Whitney $U$-test. *: significant difference. E.S.: effect size $r$.
        Note that the negative values mean that the naive architecture or JK was better than MLAP.
    }
    \TblLab{results-stat}
\end{table*}


\begin{figure*}[t]
    \centering
    \includegraphics{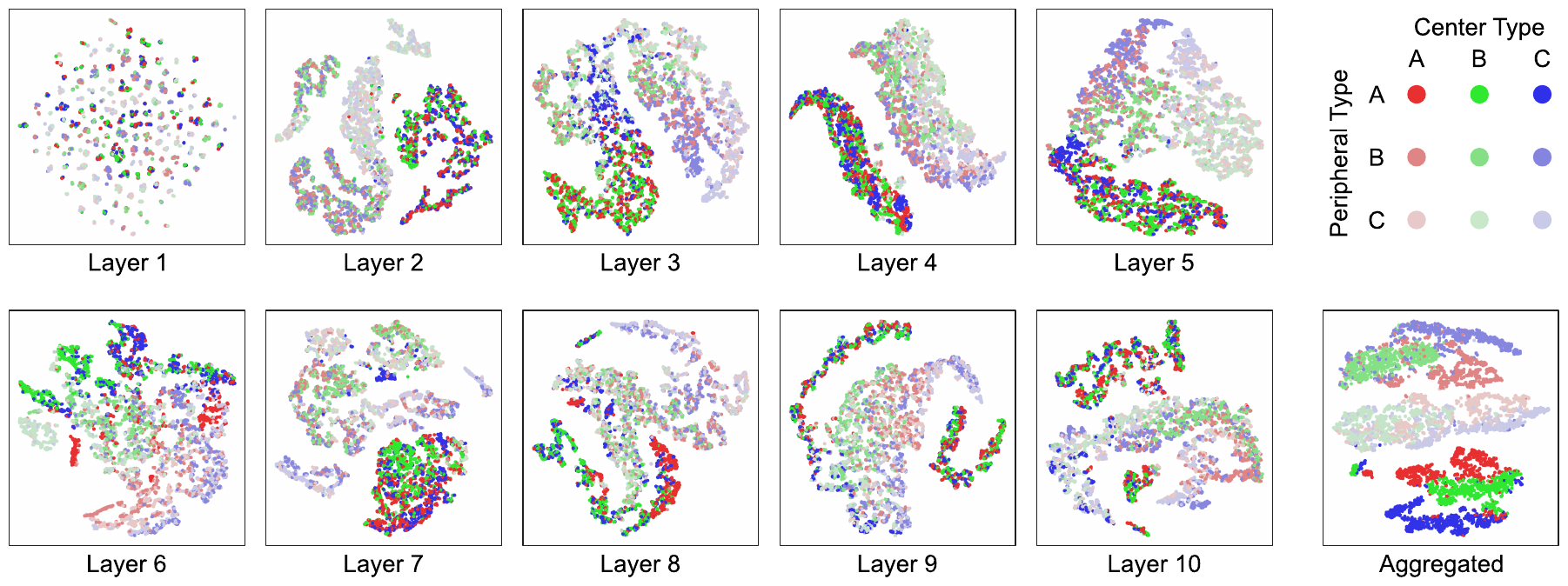}
    \caption{
        The layer-wise graph representations for the graphs in the synthetic dataset.
        They are visualized in two-dimensional spaces using t-SNE.
        Dots in each color represent samples in a class.
        (For interpretation of the reference to color in this figure legend, the reader is referred to the web version of this article.)
    }
    \FigLab{results-fractal-tsne}
\end{figure*}
\begin{figure}[t]
    \centering
    \includegraphics{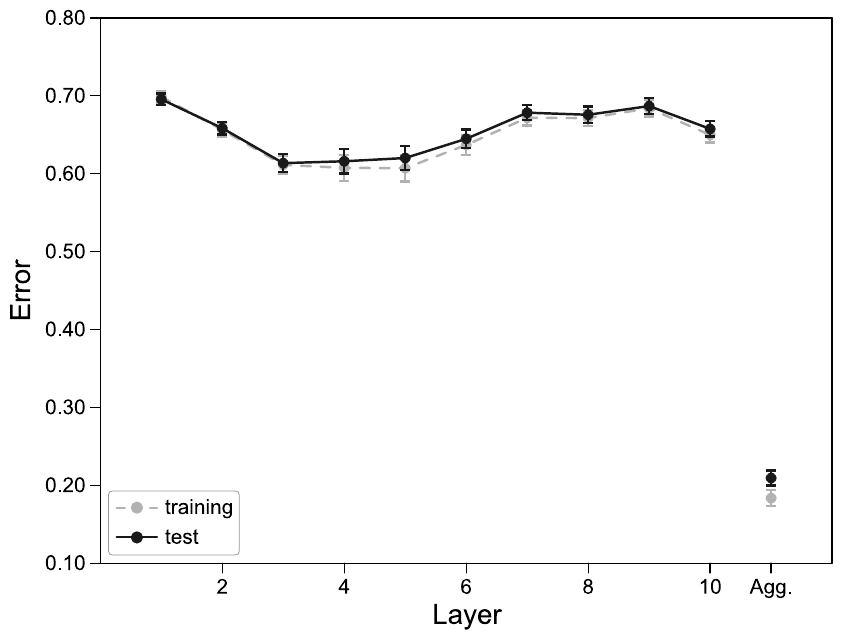}
    \caption{
        The \textcolor{red}{training and test} classification performances of the layer-wise representations computed for the graphs in the synthetic dataset.
        The ``Agg.'' in the horizontal axis indicates the classifier's performance trained with the graph representations after MLAP aggregation.
    }
    \FigLab{results-fractal-classifier}
\end{figure}
\begin{figure}[t]
    \centering
    \includegraphics{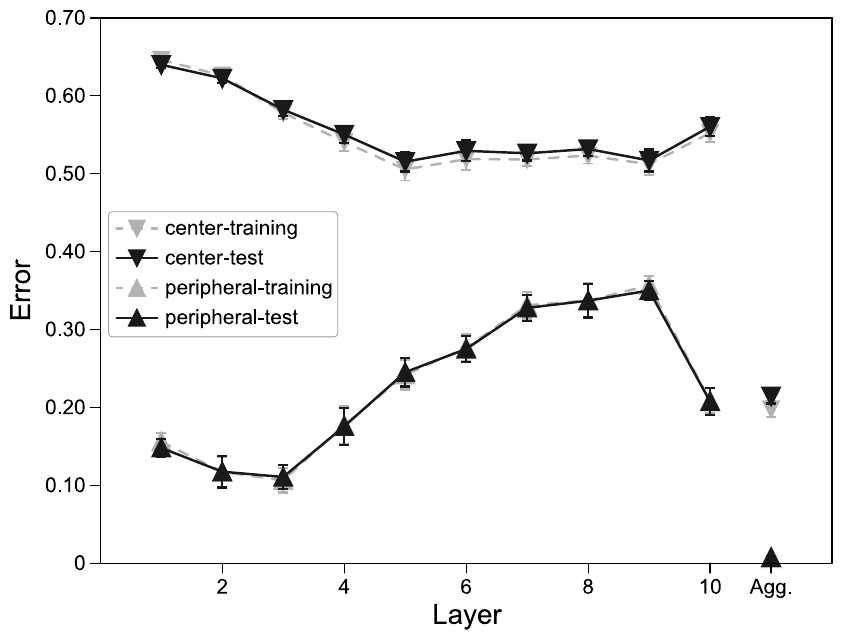}
    \caption{
        The \textcolor{red}{training and test} classification performances of the layer-wise representations computed for the graphs in the synthetic dataset.
        Instead of the 9-class classification, they are trained independently for the center type (three classes) or the peripheral types (three classes).
        The ``Agg.'' in the horizontal axis indicates the classifier's performance trained with the graph representations after MLAP aggregation.
    }
    \FigLab{results-fractal-classifier-3}
\end{figure}


\begin{figure*}[t]
    \centering
    \includegraphics{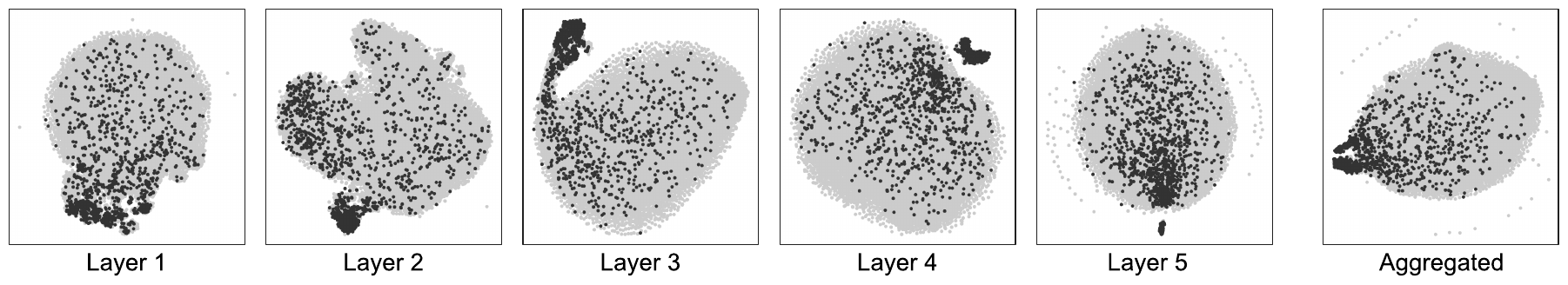}
    \caption{
        The layer-wise graph representations for ogbg-molhiv graphs.
        They are visualized in two-dimensional spaces using t-SNE.
        Each gray dot represents a \textit{negative} sample, while each black dot represents a \textit{positive} sample.
    }
    \FigLab{results-molhiv-tsne}
\end{figure*}
\begin{figure}[t]
    \centering
    \includegraphics{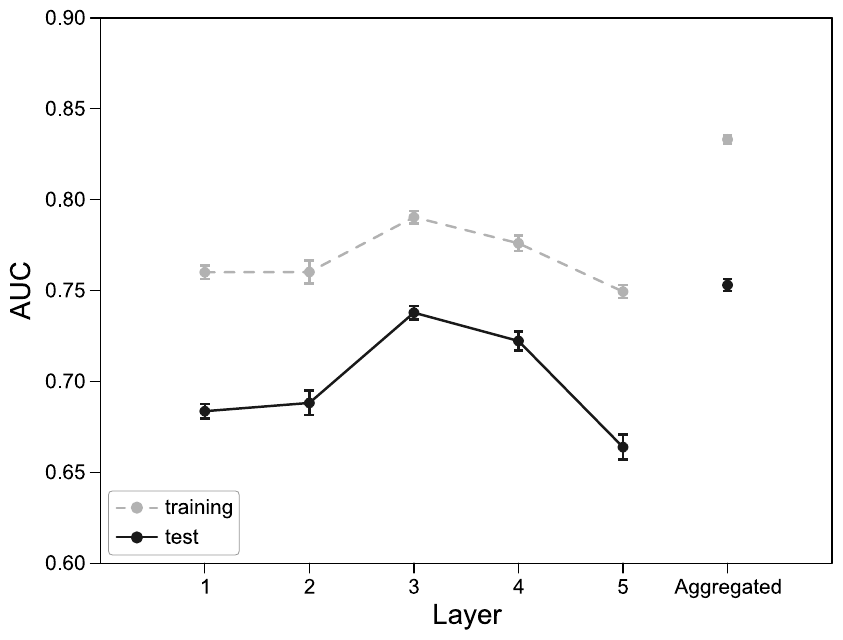}
    \caption{
        The \textcolor{red}{training and test} classification performances of the layer-wise representations computed for ogbg-molhiv graphs.
        The ``Aggregated'' in the horizontal axis indicates the classifier's performance trained with the graph representations after MLAP aggregation.
    }
    \FigLab{results-molhiv-classifier}
\end{figure}


\begin{figure*}[t]
    \centering
    \includegraphics{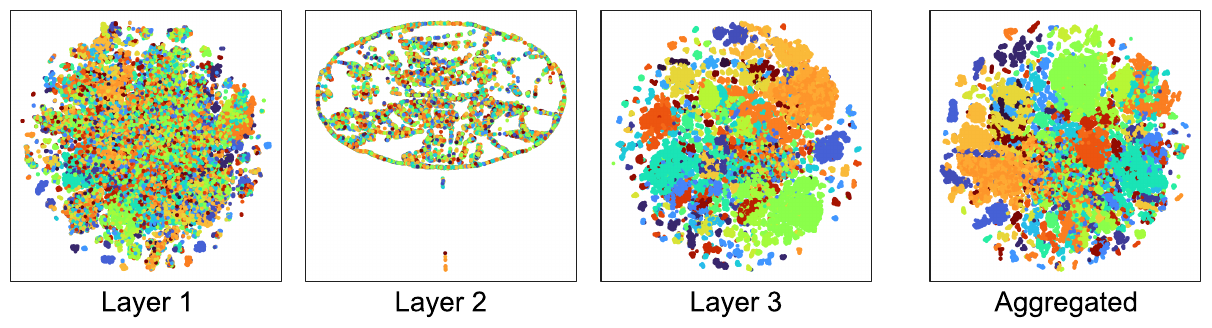}
    \caption{
        The layer-wise graph representations for ogbg-ppa graphs.
        They are visualized in two-dimensional spaces using t-SNE.
        Dots in each color represent samples in a class.
        (For interpretation of the reference to color in this figure legend, the reader is referred to the web version of this article.)
    }
    \FigLab{results-ppa-tsne}
\end{figure*}
\begin{figure}[t]
    \centering
    \includegraphics{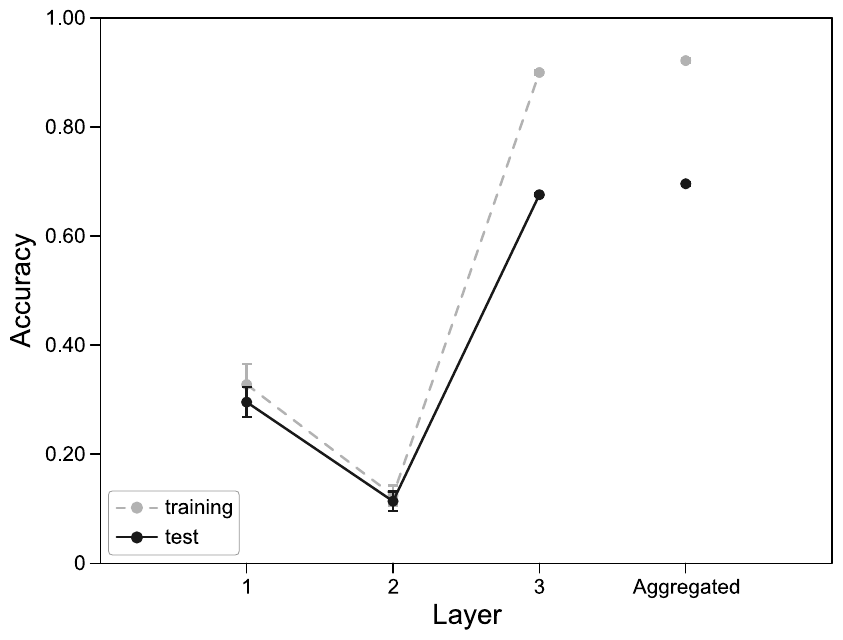}
    \caption{
        The \textcolor{red}{training and test} classification performances of the layer-wise representations computed for ogbg-ppa graphs.
        The ``Aggregated'' in the horizontal axis indicates the classifier's performance trained with the graph representations after MLAP aggregation.
    }
    \FigLab{results-ppa-classifier}
\end{figure}


\begin{figure*}[t]
    \centering
    \includegraphics{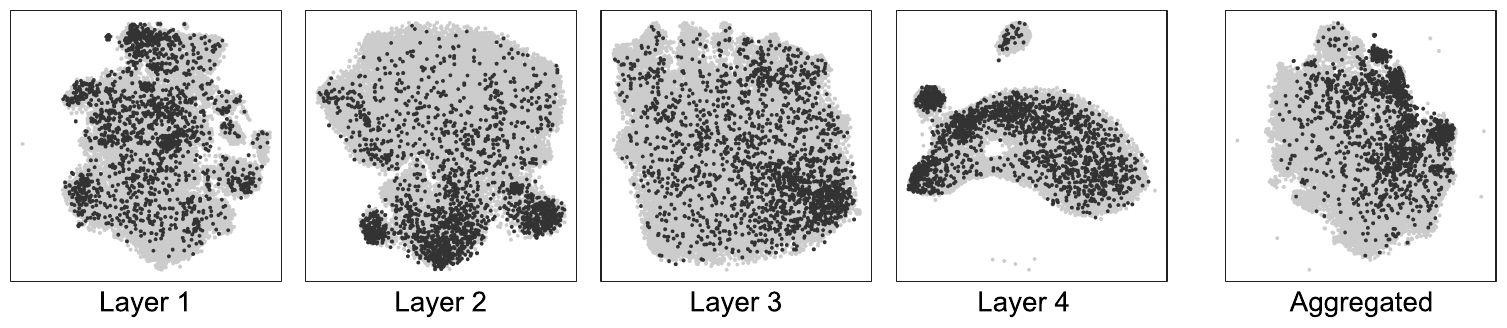}
    \caption{
        The layer-wise graph representations for MCF-7 graphs.
        They are visualized in two-dimensional spaces using t-SNE.
        Each gray dot represents a \textit{negative} sample, while each black dot represents a \textit{positive} sample.
    }
    \FigLab{results-mcf-tsne}
\end{figure*}
\begin{figure}[t]
    \centering
    \includegraphics{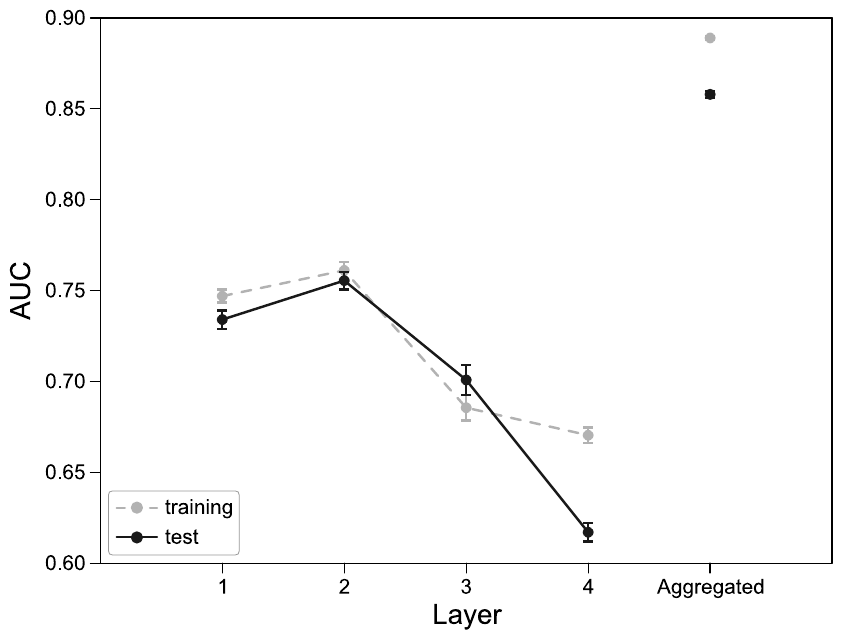}
    \caption{
        The \textcolor{red}{training and test} classification performances of the layer-wise representations computed for MCF-7 graphs.
        The ``Aggregated'' in the horizontal axis indicates the classifier's performance trained with the graph representations after MLAP aggregation.
    }
    \FigLab{results-mcf-classifier}
\end{figure}


\subsection{Analyses on Layer-Wise Representations (RQ2)}

We analyzed the layer-wise graph representations to investigate the effectiveness of the MLAP architecture in learning discriminative graph representations.
First, we computed the layer-wise graph representations and the final graph representation after MLAP aggregation for each graph in the datasets.
We conducted two different analyses on these embedded representations.

\subsubsection{t-SNE Visualization}
We visualized the distribution of those representations in a two-dimensional space using t-SNE \citep{vandermaaten2008visualizing}.
The t-SNE hyperparameters were as follows: the learning rate was 50, the number of iterations was 3000, and the perplexity was 20.

\subsubsection{Training Layer-Wise Classifiers}
We trained layer-wise classifiers to evaluate the \textit{goodness} of the layer-wise representations quantitatively.
We followed the classifier implementations in Eqs.~\EqRefBare{classifier-multi} and \EqRefBare{classifier-molhiv}, but the graph representation terms $\MlapGraphEmb{\MlapGraph}$ in those equations were replaced by the layer-wise representations $\MlapLayerEmb{\MlapGraph}{\MlapLayer}$.
These classifiers were trained on the representations of the training set.
The classification performances were tested against the representations of the test set.
The classifiers were optimized by the Adam optimizer for 30 epochs with setting the learning rate to $10^{-3}$.

\section{Results\SecLab{results}}


\subsection{Model Performances (RQ1)}

We first selected the best model with regard to the depth, the type of aggregator, and \textcolor{red}{GraphNorm (+) or ($-$)} based on the validation performance summarized in Figs.~\FigRefBare{results-fractal-depth-perf}--\textcolor{red}{\FigRefBare{results-mcf-depth-perf}}\footnote{
    For legibility, we only plotted the results of naive architecture, the best one among four JK architectures, and the best one between two MLAP architectures in Figs. \FigRefBare{results-fractal-depth-perf}--\textcolor{red}{\FigRefBare{results-mcf-depth-perf}}.
    The full results are available in \SecRefBare{app-a}.
}.
Then, we evaluated the selected models' performances using the test set (\TblRef{results-test}) and performed statistical analyses (\TblRef{results-stat}).

\subsubsection{Synthetic Dataset}

The \textit{Synthetic} column of \TblRef{results-test} summarizes the test performance in the synthetic dataset experiments.
The 9-layer MLAP-Sum \textcolor{red}{with GraphNorm} model performed the best ($0.0150\pm0.0006$).
It was better than the best performance of the baseline models: $0.0163\pm0.0005$ for 10-layer JK-LSTM-Attention \textcolor{red}{with GraphNorm}.
That is, the error rate was decreased by 8.4\%.
The statistical tests showed that MLAP performed significantly better than both the naive and the JK architectures (\TblRef{results-stat}).
The effect sizes (0.345 and 0.226) were considered as moderate to small, according to the classification given in \citet[Section~3.2]{cohen1988statistical}.

\subsubsection{ogbg-molhiv}

The \textit{ogbg-molhiv} column of \TblRef{results-test} summarizes the test performance in the ogbg-molhiv experiments.
The best performance was achieved by the 2-layer JK-Sum \textcolor{red}{with GraphNorm} model ($0.7708\pm0.0030$).
The performance of the JK model was better than the best MLAP model (6-layer MLAP-Sum \textcolor{red}{with GraphNorm}, $0.7651\pm0.0027$), but the difference was not statistically significant and the effect size was small (0.153).
MLAP performed significantly better than the naive model (2-layer \textcolor{red}{with GraphNorm}, $0.7567\pm0.0034$).

\subsubsection{ogbg-ppa}

The \textit{ogbg-ppa} column of \TblRef{results-test} summarizes the test performance in the ogbg-ppa experiments.
The best performance was $0.7198\pm0.0013$ (1-layer JK-LSTM-Attention \textcolor{red}{with GraphNorm}).
\textcolor{red}{
Note that, for this dataset, the single-layer model performed the best within each architecture; hence six out of seven types of models (naive, JK-Sum/-Concat/-MaxPool, and MLAP-Sum/-Weighted) take exactly same form.
Only JK-LSTM-Attention has extra parameters, and thus it is reasonable that it achieved the best performance thanks to these parameters, but not the hierarchical graph representations.
}

\subsubsection{MCF-7}

The \textit{MCF-7} column of \TblRef{results-test} summarizes the test performance in the MCF-7 experiments.
MLAP-Sum achieved the best test performance ($L$=5 \textcolor{red}{with GraphNorm}, $0.8634\pm0.0011$).
These performances are significantly better than the baseline performances with moderate to large effect sizes (0.407--0.\textcolor{red}{418}).


\subsection{Analyses on Layer-Wise Representations (RQ2)\protect\textcolor{red}{\protect\footnote{\textcolor{red}{In these analyses, we used GraphNorm ($-$) models so that we can avoid the interaction between MLAP and GraphNorm.}}}}

\subsubsection{Synthetic Dataset}
We visualized the learned layer-wise and the aggregated graph representations by a 10-layer MLAP-Sum model, whose validation error rate was 0.9056 (\FigRef{results-fractal-tsne}).
There were $3\times3=9$ classes of graphs in this dataset, determined by the combination of the center component type and the peripheral component type (top-right panel in \FigRef{results-fractal-tsne}).
The representations in the lower layers were highly discriminative for the \textit{peripheral} types shown by the brightness of the dots.
For the \textit{center} types shown by the hue (\IE red, green, and blue), the higher layers (Layer 6--8) seemed to have slightly better discriminative representation than the lower layers, although it was not as clear as \textit{peripheral} types.
The aggregated representations were clearly discriminative for both the center and the peripheral types.

We quantitatively evaluated this observation using layer-wise classifiers for all trained 10-layer models with 30 different random seeds.
\FigRef{results-fractal-classifier} shows the layer-wise classification performance on the training and the test sets.
Although the test error rate for each layer-wise representation was not under 0.60, the aggregated representation by MLAP achieved a significantly smaller error rate ($0.2093\pm0.0096$; $U$-test, $p<10^{-5}$ [Bonferroni corrected]).

In addition to the 9-class classifiers, \FigRef{results-fractal-classifier-3} shows the layer-wise classification performance under the \textit{3-class} settings---each classifier was trained to predict either the center type or the peripheral type.
The results in \FigRef{results-fractal-classifier-3} show the discriminability among three \textit{peripheral} types had the peak at Layer 1--3, and aggregating those layer-wise representations resulted in an error rate of almost 0.
On the other hand, the discriminability among \textit{center} types was better in higher layers (Layer 5--9), but the layer-wise error rate (best: 0.5152) was worse than that of \textit{peripheral} types.
Still, the aggregated representation achieved an error rate of 0.2142, which was much better than any of the layer-wise representations.
The 9-class classification performance (\FigRef{results-fractal-classifier}) had its peak in middle layers (Layer 3--5), which was right in between the two 3-class classifiers.
These results are consistent with the qualitative observation in \FigRef{results-fractal-tsne}, indicating graph structures in different level of locality were captured in different MLAP layers.

\subsubsection{ogbg-molhiv}
In \FigRef{results-molhiv-tsne}, We visualized the layer-wise representations by a 5-layer MLAP-Weighted model trained with the ogbg-molhiv dataset, whose validation AUC score was 0.8282.
Each gray dot represents a \textit{negative} sample, while each black dot represents a \textit{positive} sample.
The discriminability between the two classes was slightly better in the lower layers.
Aggregating those representations by taking a weighted sum produced a more localized sample distribution than any representations in the intermediate layers.

The analysis using the layer-wise classifiers supported the intuition obtained from the t-SNE visualization.
\FigRef{results-molhiv-classifier} shows the training and test AUC scores for each layer-wise classifier.
The best test score among the intermediate layers ($0.7378\pm0.0038$) was marked at Layer 3, and the score after MLAP aggregation was better than it ($0.7530\pm0.0031$).
The differences in discriminability between each layer-wise representation and the aggregated representation were significant ($U$-test, $p=0.012$ between Layer 3 and aggregated, $p<10^{-5}$ for other layers [Bonferroni corrected]) with moderate to large effect sizes ($r=0.366$--0.855).

\subsubsection{ogbg-ppa}
\FigRef{results-ppa-tsne} shows the t-SNE visualization results of the layer-wise representation by a 3-layer MLAP-Sum model ($\mathrm{Accuracy}=0.6854$).
Layer 3 showed the best discriminative representation, while representations in Layer 1 and 2 did not seem clearly discriminative.
Also, the discriminability in the MLAP-aggregated representation seemed at a similar level to Layer 3.

The layer-wise classifier analysis also showed similar results (\FigRef{results-ppa-classifier}).
The representations in Layer 3 achieved the best test score ($0.6758\pm0.0029$).
The score for the aggregated representations was slightly better ($0.6952\pm0.0029$),
whereas the differences in discriminability between each layer-wise representation and the aggregated representation were significant ($U$-test, $p=0.001$ between Layer 3 and aggregated, $p<10^{-3}$ for other layers [Bonferroni corrected]) with large effect sizes ($r=0.761$--0.845).

\subsubsection{MCF-7}
\FigRef{results-mcf-tsne} visualizes the layer-wise representation by a 4-layer MLAP-Sum model ($\mathrm{AUC}=0.8437$).
Layer 2 showed the best discriminative representation among layers 1--4, and the discriminability in the aggregated representation was even better.

The layer-wise classifier analysis supported the qualitative results.
The test performance of the layer 2 and the aggregated representations were $0.7295\pm0.0040$ and $0.8149\pm0.0016$, respectively, and the latter was significantly better than any layer-wise representations ($p<10^{-5}$ [Bonferroni corrected], effect size $r=0.859$).

\section{Discussion\SecLab{discussion}}

In this study, we proposed the MLAP architecture for GNNs, which introduces layer-wise attentional graph pooling layers and computes the final graph representation by unifying the layer-wise graph representations.
Experiments showed that our MLAP framework, which can utilize the structural information of graphs with multiple levels of localities, significantly improved the classification performance in \textcolor{red}{two out of four tested} datasets \textcolor{red}{and it showed less inferior performances to baseline methods in other two datasets}.
The performance of the \textit{naive} architecture \textcolor{red}{tended to} degrade as the number of layers increased.
This is because the deep \textit{naive} models lost the local structural information through many message passing steps due to oversmoothing\textcolor{red}{, even though the GraphNorm might mediate the effect of oversmoothing}.
On the other hand, the difference in performance between \textit{MLAP} and \textit{JK} would be because of the operation order between the graph pooling and the information aggregation from multiple levels of localities.
\textit{MLAP} computes the graph representations by $\MlapAggFn \left( \MlapLayerPool{\MlapLayer}(\MlapNodeEmb{\MlapNode}{\MlapLayer}) \right)$, whereas \textit{JK} computes them by $\MlapPool \left( \MlapJK(\MlapNodeEmb{\MlapNode}{\MlapLayer}) \right)$.
Since \textit{JK} aggregates the node representations from multiple levels of localities \textit{before} the pooling, it might be difficult for the attention mechanism to learn which node to focus on.
That is, structural information in a specific locality might be squashed before the pooling operation.
In contrast, the \textit{MLAP} architecture can tune the attention on nodes specifically in each information locality because it preserves the representations in each locality independently.
It is also supported by the observation that, for datasets with hierarchical nature, \textit{MLAP} performed better than \textit{JK} even if \textit{JK} has an aggregator with high expressivity like \textit{LSTM-Attention}\footnote{In addition, it is provable that MLAP encompasses JK under certain conditions. See \SecRefBare{app-b} for the proof.}.

The analyses on the layer-wise graph representations supported our motivation behind MLAP---GNN performance can be improved by aggregating representations in different levels of localities.
In the analyses using the synthetic dataset, the discriminability of the representations in the higher layers were worse than those in the lower layers (\FigRef{results-fractal-classifier}).
However, using 3-class classifier analyses, we showed that the learned representations had better discriminability of the \textit{peripheral} types in the lower layers, whereas the discriminability of the \textit{center} type was better in higher layers.
These results indicated that, even though the apparent classification performances in higher layers were low, those layers indeed had essential information to classify the graphs in the dataset correctly.
Aggregating layer-wise representations from multiple steps of message passing has the potential to reflect all the necessary information from various levels of localities in the final graph representation, leading to performance improvement.
The results from real-world dataset experiments were also supportive.
For the molecular datasets (ogbg-molhiv and MCF-7), the performance improvement by MLAP would be because of the hierarchical structure of biochemical molecules, whose function is determined by the combination of commonly observed substructures like carbohydrate chains and amino groups.
The MLAP architecture would effectively capture such patterns in lower layers and their combinations in higher layers.
Similarly, MLAP also worked for ogbg-ppa datasets \textcolor{red}{when the model does not have GraphNorm}. It might be because protein-protein association graphs have fractal characteristics \citep{kim2007fractality}, for which aggregating multi-locality features would be beneficial.
These results imply MLAP can utilize the compositional nature of the graphs---the feature of a whole graph is determined based on the combination of the features of smaller subgraphs.

An advantage of the aggregation mechanism of the layer-wise representations (\IE both JK and MLAP) is that such a mechanism can coincide with almost any kind of other GNN techniques.
For example, one can apply JK or MLAP for any backbone GNN architecture (\EG GCN, GIN, GAT).
Also, they can co-exist with the residual connection architectures or normalization techniques as well.
The aggregation mechanism potentially improves the performance of GNN models coordinately with these techniques.
Actually, multiple prior GRL studies have adopted JK architecture in their models and reported performance improvement.
In this study, we followed the idea to aggregate layer-wise representations, and we showed that combining the aggregation mechanism with layer-wise attention pooling can further improve the learned graph representation for graph-level classification tasks.
Our experimental results validated that MLAP can be used with GraphNorm \citep{cai2020graphnorm}.
The performance of \textit{MLAP + GraphNorm} was significantly better than \textit{naive + GraphNorm} and \textit{JK + GraphNorm} for the synthetic and MCF-7 dataset, although it was comparable to the baselines for the OGB datasets.
Comparing these results to those observed in \textit{without}-GraphNorm configuration, the advantage of MLAP over naive and JK was relatively weakened under the existence of GraphNorm.
We consider that it is because GraphNorm normalizes the node representation across the entire graph, which might prevent MLAP from learning the representation of the local structures.

Another interesting observation is that MLAP-\textit{Weighted} performed worse than MLAP-\textit{Sum} in some datasets.
We speculate that having weight parameters for layers in the aggregation process might induce instability in the training phase.
\SecRefBare{app-c} provides preliminary results supporting this hypothesis.
We will continue analyzing the cause of this phenomenon, and it might provide new insights toward further improvements in the MLAP architecture.

Designing neural network architectures by adopting knowledge in neuroscience is a popular research topic.
The multi-level attention mechanism introduced in the MLAP architecture can also be seen as an analogy of the attention mechanism in the cognitive system of humans or other primates.
Such cognitive systems, particularly the visual perception mechanism, are hierarchically organized and accompanied by hierarchical attention mechanisms.
For example, the ventral visual pathway contributes to the hierarchical computation of object recognition mechanisms \citep{kravitz2013ventral}.
In the ventral visual pathway, the neural information in the area V1 represents the raw visual inputs, and the representations are hierarchically abstracted towards the inferior temporal cortices as the receptive field (\IE locality) of the information is expanded.
\citet{deweerd1999loss} found that lesions in the cortical areas V4 and TEO, both of which are components in the ventral pathway, contribute to the attentional processing in receptive fields with different sizes.
As an example of artificial neural network studies inspired by these neuroscience studies, \citet{taylor2009hierarchical} proposed a method to autonomously control a robot using a neural network model with a hierarchical attention system, in which goal-oriented attention signals mediate the behavior of the network.
Brain-inspired neural network architectures would improve the performance or the efficiency of the models, whereas the computational studies on neural networks might contribute back to neuroscience research.
Hence, neuroscience and artificial neural networks will keep on affecting mutually and developing along with each other.

There are several possible directions to further extend the proposed methods.
First, exploring other aggregator functions than those proposed in this study, \IE \textit{Sum} and \textit{Weighted}, is needed.
For example, it is possible to design an aggregator that models the relationships among layer-wise representations, whereas the proposed aggregators treated the layer-wise representations as independent of each other.
Also, one can design an aggregator that only uses the representations in a subset of layers to reduce the computational cost, although the proposed aggregators required the layer-wise representations in all of the GNN layers.
Second, multi-stage training of the models with MLAP architecture would be beneficial.
Instead of training the entire GNN models with MLAP at once, as we did in this study, one can train the GNN backbone without MLAP first \textit{and then} fine-tune the model with the MLAP.
This kind of multi-stage training would stabilize the learning process, particularly when using the MLAP with an aggregator that has additional trainable parameters, like the \textit{MLAP-Weighted} architecture.
Lastly, our MLAP architecture would be adopted to arbitrary deep learning models, even not limited to GNNs.
For example, convolutional neural networks (CNNs) for computer vision would be good candidates.
Some CNN studies, such as U-Net \citep{ronneberger2015unet}, have already considered the hierarchy of the information processed in the neural networks.
Adopting the hierarchical attention mechanism to such models might improve their performance.

\section{Conclusion\SecLab{conclusion}}

In this study, we proposed the MLAP architecture for GNN models that aggregates graph representations from multiple levels of localities.
The results suggested that the proposed architecture was effective to learn graph representations with high discriminability.
There are many kinds of real-world networks whose properties are represented in the substructures with multiple levels of localities, and applying MLAP would improve the performances of GRL models for those graphs.

\section*{Conflict of Interest}
The authors declare no competing financial interests.

\section*{Acknowledgments}
We thank Y.~Ikutani for his valuable comments.
This work was supported by JSPS KAKENHI grant number 16H06569, 18K18108, 18K19821, and JP19J20669.

\bibliography{bib}

\clearpage

\appendix
\setcounter{table}{0}
\setcounter{figure}{0}

\section{Full Validation Performances of the Trained Models\SecLab{app-a}}

In Figs.~\FigRefBare{results-fractal-depth-perf}--\textcolor{red}{\FigRefBare{results-mcf-depth-perf}}, \textcolor{red}{we only plotted a part of validation performances for legibility.
Here, we provide the summary of the validation performances in \TblRef{results-perf} and} the full validation performances in Tables \TblRefBare{valid-full-fractal}--\TblRefBare{valid-full-mcf}.

\begin{table*}[b]
    \small
    \centering
    \begin{tabular}{ll|P{10em}P{10em}P{10em}P{10em}}
        GN & Architecture  & Synthetic ($L$)           & ogbg-molhiv ($L$)        & ogbg-ppa ($L$)           & MCF-7 ($L$)              \\ \hline\hline
        \multirow{7}{*}{\shortstack[l]{$(-)$}}
           & naive         & 0.5116 $\pm$ 0.0154 (4)   & 0.8090 $\pm$ 0.0014 (4)  & 0.6676 $\pm$ 0.0015 (1)  & 0.8023 $\pm$ 0.0011 (2)  \\ \cline{2-6}
           & JK-Sum        & 0.2347 $\pm$ 0.0082 (4)   & 0.7926 $\pm$ 0.0021 (4)  & 0.6681 $\pm$ 0.0018 (1)  & 0.7924 $\pm$ 0.0007 (2)  \\
           & JK-Concat.    & 0.2357 $\pm$ 0.0091 (10)  & 0.7786 $\pm$ 0.0019 (4)  & 0.6666 $\pm$ 0.0024 (1)  & 0.7893 $\pm$ 0.0009 (8)  \\
           & JK-MaxPool    & 0.2779 $\pm$ 0.0183 (4)   & 0.7791 $\pm$ 0.0019 (2)  & 0.6668 $\pm$ 0.0016 (1)  & 0.7901 $\pm$ 0.0016 (3)  \\
           & JK-LSTM-Att.  & 0.4109 $\pm$ 0.0215 (7)   & 0.7799 $\pm$ 0.0016 (8)  & 0.6667 $\pm$ 0.0015 (1)  & 0.7862 $\pm$ 0.0013 (2)  \\ \cline{2-6}
           & MLAP-Sum      & *0.1930 $\pm$ 0.0093 (10) & 0.8096 $\pm$ 0.0020 (3)  & *0.6691 $\pm$ 0.0050 (3) & *0.8269 $\pm$ 0.0018 (4) \\
           & MLAP-Weighted & 0.2836 $\pm$ 0.0174 (6)   & *0.8129 $\pm$ 0.0017 (5) & 0.6687 $\pm$ 0.0013 (1)  & 0.8081 $\pm$ 0.0013 (3)  \\ \hline\hline
        \multirow{7}{*}{\shortstack[l]{$(+)$}}
           & naive         & 0.0086 $\pm$ 0.0003 (9)   & 0.8156 $\pm$ 0.0022 (2)  & 0.6772 $\pm$ 0.0017 (1)  & 0.8616 $\pm$ 0.0014 (3)  \\ \cline{2-6}
           & JK-Sum        & 0.0096 $\pm$ 0.0004 (9)   & *0.8241 $\pm$ 0.0022 (2) & 0.6797 $\pm$ 0.0024 (1)  & 0.8692 $\pm$ 0.0012 (7)  \\
           & JK-Concat.    & 0.0094 $\pm$ 0.0005 (7)   & 0.8241 $\pm$ 0.0026 (2)  & 0.6700 $\pm$ 0.0026 (1)  & 0.8686 $\pm$ 0.0012 (5)  \\
           & JK-MaxPool    & 0.0089 $\pm$ 0.0004 (9)   & 0.8206 $\pm$ 0.0034 (3)  & 0.6760 $\pm$ 0.0021 (1)  & 0.8665 $\pm$ 0.0009 (5)  \\
           & JK-LSTM-Att.  & 0.0086 $\pm$ 0.0004 (10)  & 0.8115 $\pm$ 0.0018 (2)  & *0.6815 $\pm$ 0.0015 (1) & 0.8616 $\pm$ 0.0013 (3)  \\ \cline{2-6}
           & MLAP-Sum      & *0.0075 $\pm$ 0.0004 (9)  & 0.8221 $\pm$ 0.0023 (6)  & 0.6783 $\pm$ 0.0015 (1)  & *0.8720 $\pm$ 0.0011 (5) \\
           & MLAP-Weighted & 0.0100 $\pm$ 0.0004 (9)   & 0.8115 $\pm$ 0.0035 (2)  & 0.6776 $\pm$ 0.0018 (1)  & 0.8634 $\pm$ 0.0013 (4)  \\
    \end{tabular}
    \caption{
        The summary of the validation performances.
        Each cell shows the best performance of an architecture for a dataset in \textit{mean} $\pm$ \textit{standard error}.
        The numbers in the parentheses are the number of layers of the best performing models.
        We used these validation performances for model selection.
        GN: GraphNorm; JK-Concat.: JK-Concatenation; JK-LSTM-Att.: JK-LSTM-Attention.
    }
    \TblLab{results-perf}
\end{table*}

\begin{table*}[b]
    \centering
    \footnotesize
    \begin{tabular}{p{1.8em}p{9.5em}|P{8.5em}P{8.5em}P{8.5em}P{8.5em}P{8.5em}}
                            &                                    & \multicolumn{5}{c}{Number of Layers} \\
        \multirow{2}{*}{GN} & \multirow{2}{*}{Architecture}      & 1                            & 2                            & 3                   & 4                             & 5                            \\
                            &                                    & 6                            & 7                            & 8                   & 9                             & 10                           \\ \hline\hline
        \multirow{14}{*}{\shortstack[l]{($-$)}}
                            & \multirow{2}{*}{naive}             & 0.7426 $\pm$ 0.0008          & 0.5654 $\pm$ 0.0026          & 0.5382 $\pm$ 0.0173 & \textbf{0.5116} $\pm$ 0.0154  & 0.5186 $\pm$ 0.0164          \\
                            &                                    & 0.5300 $\pm$ 0.015           & 0.5213 $\pm$ 0.0167          & 0.5346 $\pm$ 0.0112 & 0.5342 $\pm$ 0.0140           & 0.5423 $\pm$ 0.0137          \\ \cline{2-7}

                            & \multirow{2}{*}{JK-Sum}            & 0.7418 $\pm$ 0.0007          & 0.5074 $\pm$ 0.0077          & 0.2521 $\pm$ 0.0083 & \textbf{0.2347} $\pm$ 0.0082  & 0.2838 $\pm$ 0.0184          \\
                            &                                    & 0.3278 $\pm$ 0.0183          & 0.4009 $\pm$ 0.0111          & 0.3947 $\pm$ 0.0098 & 0.3864 $\pm$ 0.0107           & 0.3769 $\pm$ 0.0084          \\ \cline{3-7}

                            & \multirow{2}{*}{JK-Concatenation}  & 0.7233 $\pm$ 0.0081          & 0.5447 $\pm$ 0.0092          & 0.3924 $\pm$ 0.0230 & 0.3329 $\pm$ 0.0192           & 0.3119 $\pm$ 0.0216          \\
                            &                                    & 0.2936 $\pm$ 0.0217          & 0.3123 $\pm$ 0.0285          & 0.2440 $\pm$ 0.0172 & 0.2436 $\pm$ 0.0149           & \textbf{0.2357} $\pm$ 0.0091 \\ \cline{3-7}

                            & \multirow{2}{*}{JK-MaxPool}        & 0.7384 $\pm$ 0.0042          & 0.5314 $\pm$ 0.0025          & 0.3304 $\pm$ 0.0074 & \textbf{0.2779} $\pm$ 0.0183  & 0.3240 $\pm$ 0.0177          \\
                            &                                    & 0.3441 $\pm$ 0.0175          & 0.3678 $\pm$ 0.0172          & 0.3969 $\pm$ 0.0132 & 0.4163 $\pm$ 0.0112           & 0.4229 $\pm$ 0.0116          \\ \cline{3-7}

                            & \multirow{2}{*}{JK-LSTM-Attention} & 0.7418 $\pm$ 0.0008          & 0.5869 $\pm$ 0.0119          & 0.4654 $\pm$ 0.0202 & 0.4601 $\pm$ 0.0198           & 0.4418 $\pm$ 0.0169          \\
                            &                                    & 0.4199 $\pm$ 0.0236          & \textbf{0.4109} $\pm$ 0.0215 & 0.4204 $\pm$ 0.0190 & 0.4126 $\pm$ 0.0185           & 0.4530 $\pm$ 0.0148          \\ \cline{2-7}

                            & \multirow{2}{*}{MLAP-Sum}          & 0.7423 $\pm$ 0.0009          & 0.5391 $\pm$ 0.0055          & 0.4564 $\pm$ 0.0150 & 0.3453 $\pm$ 0.0173           & 0.2906 $\pm$ 0.0145          \\
                            &                                    & 0.2250 $\pm$ 0.0107          & 0.2240 $\pm$ 0.0110          & 0.1953 $\pm$ 0.0093 & 0.2016 $\pm$ 0.0098           & \textbf{0.1930} $\pm$ 0.0093 \\ \cline{3-7}

                            & \multirow{2}{*}{MLAP-Weighted}     & 0.7431 $\pm$ 0.0007          & 0.5539 $\pm$ 0.0088          & 0.4821 $\pm$ 0.0181 & 0.4112 $\pm$ 0.0198           & 0.3536 $\pm$ 0.0197          \\
                            &                                    & \textbf{0.2836} $\pm$ 0.0174 & 0.3029 $\pm$ 0.0189          & 0.2896 $\pm$ 0.0143 & 0.2908 $\pm$ 0.0142           & 0.2861 $\pm$ 0.0170          \\ \hline\hline
        \multirow{14}{*}{\shortstack[l]{(+)}}
                            & \multirow{2}{*}{naive}             & 0.5846 $\pm$ 0.0017          & 0.2763 $\pm$ 0.0092          & 0.0597 $\pm$ 0.0023 & 0.0237 $\pm$ 0.0008           & 0.0148 $\pm$ 0.0004          \\
                            &                                    & 0.0106 $\pm$ 0.0005          & 0.0093 $\pm$ 0.0004          & 0.0094 $\pm$ 0.0005 & \textbf{0.0086} $\pm$ 0.0003  & 0.0087 $\pm$ 0.0005          \\ \cline{2-7}

                            & \multirow{2}{*}{JK-Sum}            & 0.5886 $\pm$ 0.0019          & 0.2549 $\pm$ 0.0054          & 0.0576 $\pm$ 0.0014 & 0.0236 $\pm$ 0.0007           & 0.0141 $\pm$ 0.0004          \\
                            &                                    & 0.0106 $\pm$ 0.0004          & 0.0109 $\pm$ 0.0004          & 0.0099 $\pm$ 0.0004 & \textbf{0.0096} $\pm$ 0.0004  & 0.0097 $\pm$ 0.0004          \\ \cline{3-7}

                            & \multirow{2}{*}{JK-Concatenation}  & 0.5880 $\pm$ 0.0032          & 0.2634 $\pm$ 0.0068          & 0.0554 $\pm$ 0.0011 & 0.0225 $\pm$ 0.0006           & 0.0172 $\pm$ 0.0005          \\
                            &                                    & 0.0114 $\pm$ 0.0004          & \textbf{0.0094} $\pm$ 0.0005 & 0.0097 $\pm$ 0.0004 & 0.0100 $\pm$ 0.0005           & 0.0106 $\pm$ 0.0004          \\ \cline{3-7}

                            & \multirow{2}{*}{JK-MaxPool}        & 0.5880 $\pm$ 0.0020          & 0.3203 $\pm$ 0.0025          & 0.0760 $\pm$ 0.0011 & 0.0250 $\pm$ 0.0007           & 0.0146 $\pm$ 0.0005          \\
                            &                                    & 0.0110 $\pm$ 0.0005          & 0.0090 $\pm$ 0.0004          & 0.0091 $\pm$ 0.0004 & \textbf{0.0089} $\pm$ 0.0004  & 0.0090 $\pm$ 0.0004          \\ \cline{3-7}

                            & \multirow{2}{*}{JK-LSTM-Attention} & 0.5874 $\pm$ 0.0019          & 0.2693 $\pm$ 0.0064          & 0.0615 $\pm$ 0.0021 & 0.0251 $\pm$ 0.0010           & 0.0144 $\pm$ 0.0004          \\
                            &                                    & 0.0100 $\pm$ 0.0005          & 0.0097 $\pm$ 0.0004          & 0.0090 $\pm$ 0.0004 & 0.0087 $\pm$ 0.0005           & \textbf{0.0086} $\pm$ 0.0004 \\ \cline{2-7}

                            & \multirow{2}{*}{MLAP-Sum}          & 0.5878 $\pm$ 0.0019          & 0.2669 $\pm$ 0.0075          & 0.0574 $\pm$ 0.0018 & 0.0220 $\pm$ 0.0011           & 0.0153 $\pm$ 0.0006          \\
                            &                                    & 0.0101 $\pm$ 0.0003          & 0.0091 $\pm$ 0.0005          & 0.0086 $\pm$ 0.0004 & \textbf{*0.0075} $\pm$ 0.0004 & 0.0085 $\pm$ 0.0004          \\ \cline{3-7}

                            & \multirow{2}{*}{MLAP-Weighted}     & 0.5868 $\pm$ 0.0018          & 0.2840 $\pm$ 0.0090          & 0.0696 $\pm$ 0.0021 & 0.0284 $\pm$ 0.0006           & 0.0161 $\pm$ 0.0006          \\
                            &                                    & 0.0126 $\pm$ 0.0005          & 0.0103 $\pm$ 0.0004          & 0.0104 $\pm$ 0.0004 & \textbf{0.0100} $\pm$ 0.0004  & 0.0100 $\pm$ 0.0005          \\
    \end{tabular}
    \caption{Full validation performances in the synthetic dataset experiments for model selection. \textbf{bold}:~best performance in \textcolor{red}{an (aggregator, GraphNorm [+/$-$]) combination}; \textbf{*}:~the overall best performance.}
    \TblLab{valid-full-fractal}
\end{table*}

\begin{table*}[b]
    \centering
    \footnotesize
    \begin{tabular}{p{1.8em}p{9.5em}|P{8.5em}P{8.5em}P{8.5em}P{8.5em}P{8.5em}}
                            &                                    & \multicolumn{5}{c}{Number of Layers} \\
        \multirow{2}{*}{GN} & \multirow{2}{*}{Architecture}      & 1                            & 2                             & 3                            & 4                            & 5                            \\
                            &                                    & 6                            & 7                             & 8                            & 9                            & 10                           \\ \hline\hline
        \multirow{14}{*}{\shortstack[l]{($-$)}}
                            & \multirow{2}{*}{naive}             & 0.7467 $\pm$ 0.0020          & 0.7731 $\pm$ 0.0028           & 0.8014 $\pm$ 0.0014          & \textbf{0.8090} $\pm$ 0.0014 & 0.8052 $\pm$ 0.0013          \\
                            &                                    & 0.7909 $\pm$ 0.0013          & 0.7820 $\pm$ 0.0013           & 0.7664 $\pm$ 0.0041          & 0.7599 $\pm$ 0.0042          & 0.7631 $\pm$ 0.0022          \\ \cline{2-7}

                            & \multirow{2}{*}{JK-Sum}            & 0.7446 $\pm$ 0.0022          & 0.7735 $\pm$ 0.0016           & 0.7885 $\pm$ 0.0016          & \textbf{0.7926} $\pm$ 0.0021 & 0.7775 $\pm$ 0.0020          \\
                            &                                    & 0.7726 $\pm$ 0.0018          & 0.7753 $\pm$ 0.0021           & 0.7738 $\pm$ 0.0017          & 0.7706 $\pm$ 0.0018          & 0.7720 $\pm$ 0.0022          \\ \cline{3-7}

                            & \multirow{2}{*}{JK-Concatenation}  & 0.7538 $\pm$ 0.0025          & 0.7677 $\pm$ 0.0028           & 0.7772 $\pm$ 0.0019          & \textbf{0.7786} $\pm$ 0.0019 & 0.7713 $\pm$ 0.0021          \\
                            &                                    & 0.7732 $\pm$ 0.0018          & 0.7731 $\pm$ 0.0019           & 0.7746 $\pm$ 0.0018          & 0.7693 $\pm$ 0.0024          & 0.7681 $\pm$ 0.0023          \\ \cline{3-7}

                            & \multirow{2}{*}{JK-MaxPool}        & 0.7436 $\pm$ 0.0018          & \textbf{0.7791} $\pm$ 0.0019  & 0.7723 $\pm$ 0.0017          & 0.7753 $\pm$ 0.0030          & 0.7689 $\pm$ 0.0020          \\
                            &                                    & 0.7667 $\pm$ 0.0025          & 0.7699 $\pm$ 0.0020           & 0.7655 $\pm$ 0.0016          & 0.7661 $\pm$ 0.0021          & 0.7667 $\pm$ 0.0021          \\ \cline{3-7}

                            & \multirow{2}{*}{JK-LSTM-Attention} & 0.7464 $\pm$ 0.0022          & 0.7619 $\pm$ 0.0022           & 0.7661 $\pm$ 0.0013          & 0.7711 $\pm$ 0.0017          & 0.7689 $\pm$ 0.0018          \\
                            &                                    & 0.7741 $\pm$ 0.0017          & 0.7735 $\pm$ 0.0015           & \textbf{0.7799} $\pm$ 0.0016 & 0.7788 $\pm$ 0.0024          & 0.7738 $\pm$ 0.0025          \\ \cline{2-7}

                            & \multirow{2}{*}{MLAP-Sum}          & 0.7452 $\pm$ 0.0023          & 0.7787 $\pm$ 0.0025           & \textbf{0.8096} $\pm$ 0.0020 & 0.8002 $\pm$ 0.0029          & 0.8003 $\pm$ 0.0024          \\
                            &                                    & 0.8019 $\pm$ 0.0023          & 0.8007 $\pm$ 0.0025           & 0.8015 $\pm$ 0.0024          & 0.7971 $\pm$ 0.0018          & 0.8029 $\pm$ 0.0025          \\ \cline{3-7}

                            & \multirow{2}{*}{MLAP-Weighted}     & 0.7457 $\pm$ 0.0024          & 0.7702 $\pm$ 0.0023           & 0.7987 $\pm$ 0.0019          & 0.8107 $\pm$ 0.0020          & \textbf{0.8129} $\pm$ 0.0017 \\
                            &                                    & 0.8076 $\pm$ 0.0024          & 0.8019 $\pm$ 0.0023           & 0.7977 $\pm$ 0.0026          & 0.8040 $\pm$ 0.0024          & 0.8027 $\pm$ 0.0020          \\ \hline\hline
        \multirow{14}{*}{\shortstack[l]{(+)}}
                            & \multirow{2}{*}{naive}             & 0.8055 $\pm$ 0.0038          & \textbf{0.8159} $\pm$ 0.0022  & 0.8157 $\pm$ 0.0022          & 0.7958 $\pm$ 0.0031          & 0.7868 $\pm$ 0.0038          \\
                            &                                    & 0.7852 $\pm$ 0.0035          & 0.7733 $\pm$ 0.0037           & 0.7736 $\pm$ 0.0035          & 0.7634 $\pm$ 0.0042          & 0.7638 $\pm$ 0.0042          \\ \cline{2-7}

                            & \multirow{2}{*}{JK-Sum}            & 0.8041 $\pm$ 0.0026          & \textbf{*0.8241} $\pm$ 0.0022 & 0.8198 $\pm$ 0.0024          & 0.8157 $\pm$ 0.0023          & 0.8144 $\pm$ 0.0022          \\
                            &                                    & 0.8120 $\pm$ 0.0026          & 0.8096 $\pm$ 0.0027           & 0.8028 $\pm$ 0.0026          & 0.8088 $\pm$ 0.0022          & 0.7994 $\pm$ 0.0023          \\ \cline{3-7}

                            & \multirow{2}{*}{JK-Concatenation}  & 0.8042 $\pm$ 0.0028          & \textbf{0.8241} $\pm$ 0.0026  & 0.8201 $\pm$ 0.0023          & 0.8212 $\pm$ 0.0028          & 0.8217 $\pm$ 0.0025          \\
                            &                                    & 0.8135 $\pm$ 0.0026          & 0.8095 $\pm$ 0.0031           & 0.8115 $\pm$ 0.0032          & 0.8099 $\pm$ 0.0027          & 0.8107 $\pm$ 0.0031          \\ \cline{3-7}

                            & \multirow{2}{*}{JK-MaxPool}        & 0.8039 $\pm$ 0.0032          & 0.8144 $\pm$ 0.0030           & \textbf{0.8206} $\pm$ 0.0034 & 0.8155 $\pm$ 0.0030          & 0.8100 $\pm$ 0.0032          \\
                            &                                    & 0.8068 $\pm$ 0.0027          & 0.8024 $\pm$ 0.0030           & 0.7979 $\pm$ 0.0030          & 0.7955 $\pm$ 0.0031          & 0.7954 $\pm$ 0.0031          \\ \cline{3-7}

                            & \multirow{2}{*}{JK-LSTM-Attention} & 0.8037 $\pm$ 0.0030          & \textbf{0.8115} $\pm$ 0.0018  & 0.8058 $\pm$ 0.0028          & 0.8043 $\pm$ 0.0034          & 0.8045 $\pm$ 0.0032          \\
                            &                                    & 0.8082 $\pm$ 0.0027          & 0.8098 $\pm$ 0.0028           & 0.7874 $\pm$ 0.0040          & 0.7924 $\pm$ 0.0039          & 0.7791 $\pm$ 0.0033          \\ \cline{2-7}

                            & \multirow{2}{*}{MLAP-Sum}          & 0.8070 $\pm$ 0.0035          & 0.8121 $\pm$ 0.0031           & 0.8149 $\pm$ 0.0027          & 0.8204 $\pm$ 0.0028          & 0.8190 $\pm$ 0.0024          \\
                            &                                    & \textbf{0.8221} $\pm$ 0.0023 & 0.8178 $\pm$ 0.0037           & 0.8166 $\pm$ 0.0019          & 0.8140 $\pm$ 0.0023          & 0.8082 $\pm$ 0.0028          \\ \cline{3-7}

                            & \multirow{2}{*}{MLAP-Weighted}     & 0.8027 $\pm$ 0.0031          & \textbf{0.8115} $\pm$ 0.0035  & 0.8065 $\pm$ 0.0029          & 0.8088 $\pm$ 0.0025          & 0.8045 $\pm$ 0.0029          \\
                            &                                    & 0.8024 $\pm$ 0.0037          & 0.8015 $\pm$ 0.0020           & 0.8036 $\pm$ 0.0028          & 0.7947 $\pm$ 0.0037          & 0.7951 $\pm$ 0.0034          \\
    \end{tabular}
    \caption{Full validation performances in the ogbg-molhiv experiments for model selection. \textbf{bold}:~best performance in \textcolor{red}{an (aggregator, GraphNorm [+/$-$]) combination}; \textbf{*}:~the overall best performance.}
    \TblLab{valid-full-molhiv}
\end{table*}

\begin{table*}[b]
    \centering
    \footnotesize
    \begin{tabular}{p{1.8em}p{9.5em}|P{8.5em}P{8.5em}P{8.5em}P{8.5em}P{8.5em}}
                            &                                    & \multicolumn{5}{c}{Number of Layers} \\
        \multirow{2}{*}{GN} & \multirow{2}{*}{Architecture}      & 1                             & 2                   & 3                            & 4                   & 5                   \\
                            &                                    & 6                             & 7                   & 8                            & 9                   & 10                  \\ \hline\hline
        \multirow{14}{*}{\shortstack[l]{($-$)}}
                            & \multirow{2}{*}{naive}             & \textbf{0.6676} $\pm$ 0.0015  & 0.6639 $\pm$ 0.0034 & 0.6622 $\pm$ 0.0048          & 0.6560 $\pm$ 0.0037 & 0.6395 $\pm$ 0.0029 \\
                            &                                    & 0.6153 $\pm$ 0.0032           & 0.5936 $\pm$ 0.0054 & 0.5746 $\pm$ 0.0060          & 0.5442 $\pm$ 0.0099 & 0.4973 $\pm$ 0.0080 \\ \cline{2-7}

                            & \multirow{2}{*}{JK-Sum}            & \textbf{0.6681} $\pm$ 0.0018  & 0.6610 $\pm$ 0.0042 & 0.6652 $\pm$ 0.0060          & 0.6569 $\pm$ 0.0052 & 0.6440 $\pm$ 0.0044 \\
                            &                                    & 0.6229 $\pm$ 0.0059           & 0.6206 $\pm$ 0.0051 & 0.6123 $\pm$ 0.0052          & 0.5913 $\pm$ 0.0050 & 0.5622 $\pm$ 0.0090 \\ \cline{3-7}

                            & \multirow{2}{*}{JK-Concatenation}  & \textbf{0.6666} $\pm$ 0.0024  & 0.6352 $\pm$ 0.0041 & 0.6636 $\pm$ 0.0100          & 0.6420 $\pm$ 0.0057 & 0.6467 $\pm$ 0.0066 \\
                            &                                    & 0.6257 $\pm$ 0.0053           & 0.6219 $\pm$ 0.0096 & 0.6152 $\pm$ 0.0074          & 0.6011 $\pm$ 0.0064 & 0.5876 $\pm$ 0.0070 \\ \cline{3-7}

                            & \multirow{2}{*}{JK-MaxPool}        & \textbf{0.6668} $\pm$ 0.0016  & 0.6076 $\pm$ 0.0031 & 0.6049 $\pm$ 0.0076          & 0.6070 $\pm$ 0.0088 & 0.5962 $\pm$ 0.0072 \\
                            &                                    & 0.6041 $\pm$ 0.0057           & 0.6023 $\pm$ 0.0061 & 0.5838 $\pm$ 0.0055          & 0.5837 $\pm$ 0.0075 & 0.5292 $\pm$ 0.0104 \\ \cline{3-7}

                            & \multirow{2}{*}{JK-LSTM-Attention} & \textbf{0.6667} $\pm$ 0.0015  & 0.6127 $\pm$ 0.0019 & 0.5910 $\pm$ 0.0069          & 0.5888 $\pm$ 0.0048 & 0.5542 $\pm$ 0.0116 \\
                            &                                    & 0.5020 $\pm$ 0.0333           & 0.4251 $\pm$ 0.0452 & 0.4237 $\pm$ 0.0479          & 0.4499 $\pm$ 0.0265 & 0.3046 $\pm$ 0.0305 \\ \cline{2-7}

                            & \multirow{2}{*}{MLAP-Sum}          & 0.6665 $\pm$ 0.0014           & 0.6568 $\pm$ 0.0042 & \textbf{0.6691} $\pm$ 0.0050 & 0.6512 $\pm$ 0.0059 & 0.6529 $\pm$ 0.0052 \\
                            &                                    & 0.6246 $\pm$ 0.0069           & 0.6144 $\pm$ 0.0068 & 0.5959 $\pm$ 0.0071          & 0.5732 $\pm$ 0.0106 & 0.5721 $\pm$ 0.0074 \\ \cline{3-7}

                            & \multirow{2}{*}{MLAP-Weighted}     & \textbf{0.6687} $\pm$ 0.0013  & 0.6463 $\pm$ 0.0034 & 0.6542 $\pm$ 0.0078          & 0.6358 $\pm$ 0.0038 & 0.6305 $\pm$ 0.0047 \\
                            &                                    & 0.6192 $\pm$ 0.0046           & 0.6023 $\pm$ 0.0079 & 0.5642 $\pm$ 0.0065          & 0.5583 $\pm$ 0.0096 & 0.5053 $\pm$ 0.0155 \\ \hline\hline
        \multirow{14}{*}{\shortstack[l]{(+)}}
                            & \multirow{2}{*}{naive}             & \textbf{0.6772} $\pm$ 0.0017  & 0.6388 $\pm$ 0.0031 & 0.6263 $\pm$ 0.0028          & 0.6379 $\pm$ 0.0022 & 0.6448 $\pm$ 0.0023 \\
                            &                                    & 0.6403 $\pm$ 0.0023           & 0.6418 $\pm$ 0.0025 & 0.6427 $\pm$ 0.0017          & 0.6402 $\pm$ 0.0025 & 0.6441 $\pm$ 0.0027 \\ \cline{2-7}

                            & \multirow{2}{*}{JK-Sum}            & \textbf{0.6797} $\pm$ 0.0024  & 0.6408 $\pm$ 0.0025 & 0.6370 $\pm$ 0.0034          & 0.6414 $\pm$ 0.0046 & 0.6415 $\pm$ 0.0045 \\
                            &                                    & 0.6291 $\pm$ 0.0020           & 0.6315 $\pm$ 0.0039 & 0.6254 $\pm$ 0.0027          & 0.6295 $\pm$ 0.0026 & 0.6274 $\pm$ 0.0032 \\ \cline{3-7}

                            & \multirow{2}{*}{JK-Concatenation}  & \textbf{0.6700} $\pm$ 0.0026  & 0.6496 $\pm$ 0.0039 & 0.6537 $\pm$ 0.0032          & 0.6517 $\pm$ 0.0040 & 0.6491 $\pm$ 0.0025 \\
                            &                                    & 0.6516 $\pm$ 0.0036           & 0.6466 $\pm$ 0.0035 & 0.6465 $\pm$ 0.0034          & 0.6479 $\pm$ 0.0032 & 0.6429 $\pm$ 0.0030 \\ \cline{3-7}

                            & \multirow{2}{*}{JK-MaxPool}        & \textbf{0.6760} $\pm$ 0.0021  & 0.6475 $\pm$ 0.0021 & 0.6415 $\pm$ 0.0029          & 0.6446 $\pm$ 0.0034 & 0.6349 $\pm$ 0.0015 \\
                            &                                    & 0.6362 $\pm$ 0.0040           & 0.6313 $\pm$ 0.0030 & 0.6316 $\pm$ 0.0039          & 0.6382 $\pm$ 0.0055 & 0.6353 $\pm$ 0.0037 \\ \cline{3-7}

                            & \multirow{2}{*}{JK-LSTM-Attention} & \textbf{*0.6815} $\pm$ 0.0015 & 0.6522 $\pm$ 0.0049 & 0.6634 $\pm$ 0.0024          & 0.6630 $\pm$ 0.0016 & 0.6630 $\pm$ 0.0035 \\
                            &                                    & 0.6629 $\pm$ 0.0020           & 0.6658 $\pm$ 0.0024 & 0.6627 $\pm$ 0.0021          & 0.6647 $\pm$ 0.0027 & 0.6642 $\pm$ 0.0025 \\ \cline{2-7}

                            & \multirow{2}{*}{MLAP-Sum}          & \textbf{0.6783} $\pm$ 0.0015  & 0.6373 $\pm$ 0.0031 & 0.6362 $\pm$ 0.0030          & 0.6370 $\pm$ 0.0022 & 0.6353 $\pm$ 0.0030 \\
                            &                                    & 0.6377 $\pm$ 0.0042           & 0.6326 $\pm$ 0.0026 & 0.6336 $\pm$ 0.0026          & 0.6359 $\pm$ 0.0025 & 0.6377 $\pm$ 0.0036 \\ \cline{3-7}

                            & \multirow{2}{*}{MLAP-Weighted}     & \textbf{0.6776} $\pm$ 0.0018  & 0.6327 $\pm$ 0.0029 & 0.6538 $\pm$ 0.0017          & 0.6550 $\pm$ 0.0045 & 0.6475 $\pm$ 0.0036 \\
                            &                                    & 0.6561 $\pm$ 0.0048           & 0.6598 $\pm$ 0.0048 & 0.6535 $\pm$ 0.0077          & 0.6443 $\pm$ 0.0055 & 0.6503 $\pm$ 0.0045 \\
    \end{tabular}
    \caption{Full validation performances in the ogbg-ppa experiments for model selection. \textbf{bold}:~best performance in \textcolor{red}{an (aggregator, GraphNorm [+/$-$]) combination}; \textbf{*}:~the overall best performance.}
    \TblLab{valid-full-ppa}
\end{table*}

\begin{table*}[b]
    \centering
    \footnotesize
    \begin{tabular}{p{1.8em}p{9.5em}|P{8.5em}P{8.5em}P{8.5em}P{8.5em}P{8.5em}}
                            &                                    & \multicolumn{5}{c}{Number of Layers} \\
        \multirow{2}{*}{GN} & \multirow{2}{*}{Architecture}      & 1                   & 2                            & 3                            & 4                            & 5                             \\
                            &                                    & 6                   & 7                            & 8                            & 9                            & 10                            \\ \hline\hline
        \multirow{14}{*}{\shortstack[l]{($-$)}}
                            & \multirow{2}{*}{naive}             & 0.7653 $\pm$ 0.0010 & \textbf{0.8023} $\pm$ 0.0011 & 0.7960 $\pm$ 0.0024          & 0.7996 $\pm$ 0.0011          & 0.7968 $\pm$ 0.0012           \\
                            &                                    & 0.7689 $\pm$ 0.0033 & 0.7361 $\pm$ 0.0045          & 0.7101 $\pm$ 0.0045          & 0.7086 $\pm$ 0.0035          & 0.7097 $\pm$ 0.0049           \\ \cline{2-7}

                            & \multirow{2}{*}{JK-Sum}            & 0.7641 $\pm$ 0.0011 & \textbf{0.7924} $\pm$ 0.0007 & 0.7836 $\pm$ 0.0015          & 0.7849 $\pm$ 0.0007          & 0.7884 $\pm$ 0.0013           \\
                            &                                    & 0.7866 $\pm$ 0.0017 & 0.7853 $\pm$ 0.0013          & 0.7846 $\pm$ 0.0010          & 0.7859 $\pm$ 0.0016          & 0.7827 $\pm$ 0.0011           \\ \cline{3-7}

                            & \multirow{2}{*}{JK-Concatenation}  & 0.7648 $\pm$ 0.0010 & 0.7868 $\pm$ 0.0008          & 0.7797 $\pm$ 0.0013          & 0.7857 $\pm$ 0.0007          & 0.7862 $\pm$ 0.0011           \\
                            &                                    & 0.7884 $\pm$ 0.0016 & 0.7875 $\pm$ 0.0011          & \textbf{0.7893} $\pm$ 0.0009 & 0.7890 $\pm$ 0.0010          & 0.7881 $\pm$ 0.0011           \\ \cline{3-7}

                            & \multirow{2}{*}{JK-MaxPool}        & 0.7639 $\pm$ 0.0011 & 0.7860 $\pm$ 0.0009          & \textbf{0.7901} $\pm$ 0.0016 & 0.7811 $\pm$ 0.0007          & 0.7825 $\pm$ 0.0010           \\
                            &                                    & 0.7814 $\pm$ 0.0009 & 0.7802 $\pm$ 0.0009          & 0.7821 $\pm$ 0.0012          & 0.7808 $\pm$ 0.0010          & 0.7798 $\pm$ 0.0011           \\ \cline{3-7}

                            & \multirow{2}{*}{JK-LSTM-Attention} & 0.7651 $\pm$ 0.0013 & \textbf{0.7862} $\pm$ 0.0013 & 0.7794 $\pm$ 0.0009          & 0.7780 $\pm$ 0.0012          & 0.7770 $\pm$ 0.0014           \\
                            &                                    & 0.7814 $\pm$ 0.0014 & 0.7787 $\pm$ 0.0010          & 0.7784 $\pm$ 0.0011          & 0.7775 $\pm$ 0.0015          & 0.7808 $\pm$ 0.0016           \\ \cline{2-7}

                            & \multirow{2}{*}{MLAP-Sum}          & 0.7642 $\pm$ 0.0011 & 0.7790 $\pm$ 0.0011          & 0.8138 $\pm$ 0.0010          & \textbf{0.8269} $\pm$ 0.0018 & 0.8243 $\pm$ 0.0014           \\
                            &                                    & 0.8216 $\pm$ 0.0014 & 0.8202 $\pm$ 0.0015          & 0.8200 $\pm$ 0.0012          & 0.8228 $\pm$ 0.0012          & 0.8223 $\pm$ 0.0012           \\ \cline{3-7}

                            & \multirow{2}{*}{MLAP-Weighted}     & 0.7640 $\pm$ 0.0010 & 0.7738 $\pm$ 0.0012          & \textbf{0.8081} $\pm$ 0.0013 & 0.8002 $\pm$ 0.0012          & 0.8010 $\pm$ 0.0013           \\
                            &                                    & 0.8029 $\pm$ 0.0015 & 0.7996 $\pm$ 0.0022          & 0.7985 $\pm$ 0.0016          & 0.8019 $\pm$ 0.0016          & 0.7996 $\pm$ 0.0018           \\ \hline\hline
        \multirow{14}{*}{\shortstack[l]{(+)}}
                            & \multirow{2}{*}{naive}             & 0.8271 $\pm$ 0.0015 & 0.8536 $\pm$ 0.0013          & \textbf{0.8616} $\pm$ 0.0014 & 0.8613 $\pm$ 0.0016          & 0.8574 $\pm$ 0.0020           \\
                            &                                    & 0.8510 $\pm$ 0.0024 & 0.8438 $\pm$ 0.0021          & 0.8308 $\pm$ 0.0024          & 0.8265 $\pm$ 0.0027          & 0.8108 $\pm$ 0.0019           \\ \cline{2-7}

                            & \multirow{2}{*}{JK-Sum}            & 0.8259 $\pm$ 0.0013 & 0.8564 $\pm$ 0.0010          & 0.8635 $\pm$ 0.0011          & 0.8650 $\pm$ 0.0012          & 0.8671 $\pm$ 0.0010           \\
                            &                                    & 0.8664 $\pm$ 0.0015 & \textbf{0.8692} $\pm$ 0.0012 & 0.8611 $\pm$ 0.0022          & 0.8623 $\pm$ 0.0020          & 0.8611 $\pm$ 0.0025           \\ \cline{3-7}

                            & \multirow{2}{*}{JK-Concatenation}  & 0.8301 $\pm$ 0.0016 & 0.8598 $\pm$ 0.0011          & 0.8664 $\pm$ 0.0011          & 0.8684 $\pm$ 0.0011          & \textbf{0.8686} $\pm$ 0.0012  \\
                            &                                    & 0.8653 $\pm$ 0.0010 & 0.8613 $\pm$ 0.0010          & 0.8606 $\pm$ 0.0015          & 0.8563 $\pm$ 0.0013          & 0.8518 $\pm$ 0.0013           \\ \cline{3-7}

                            & \multirow{2}{*}{JK-MaxPool}        & 0.8267 $\pm$ 0.0010 & 0.8558 $\pm$ 0.0012          & 0.8633 $\pm$ 0.0013          & 0.8622 $\pm$ 0.0013          & \textbf{0.8665} $\pm$ 0.0009  \\
                            &                                    & 0.8629 $\pm$ 0.0014 & 0.8619 $\pm$ 0.0012          & 0.8583 $\pm$ 0.0019          & 0.8541 $\pm$ 0.0016          & 0.8484 $\pm$ 0.0019           \\ \cline{3-7}

                            & \multirow{2}{*}{JK-LSTM-Attention} & 0.8257 $\pm$ 0.0015 & 0.8552 $\pm$ 0.0010          & \textbf{0.8616} $\pm$ 0.0013 & 0.8578 $\pm$ 0.0014          & 0.8588 $\pm$ 0.0015           \\
                            &                                    & 0.8572 $\pm$ 0.0019 & 0.8519 $\pm$ 0.0024          & 0.8469 $\pm$ 0.0027          & 0.8375 $\pm$ 0.0027          & 0.8369 $\pm$ 0.0024           \\ \cline{2-7}

                            & \multirow{2}{*}{MLAP-Sum}          & 0.8267 $\pm$ 0.0012 & 0.8583 $\pm$ 0.0010          & 0.8653 $\pm$ 0.0010          & 0.8686 $\pm$ 0.0011          & \textbf{*0.8720} $\pm$ 0.0011 \\
                            &                                    & 0.8695 $\pm$ 0.0011 & 0.8680 $\pm$ 0.0013          & 0.8668 $\pm$ 0.0012          & 0.8630 $\pm$ 0.0021          & 0.8634 $\pm$ 0.0014           \\ \cline{3-7}

                            & \multirow{2}{*}{MLAP-Weighted}     & 0.8270 $\pm$ 0.0011 & 0.8555 $\pm$ 0.0013          & 0.8626 $\pm$ 0.0011          & \textbf{0.8634} $\pm$ 0.0013 & 0.8605 $\pm$ 0.0008           \\
                            &                                    & 0.8552 $\pm$ 0.0013 & 0.8559 $\pm$ 0.0010          & 0.8497 $\pm$ 0.0015          & 0.8413 $\pm$ 0.0018          & 0.8357 $\pm$ 0.0017           \\
    \end{tabular}
    \caption{Full results in the MCF-7 dataset experiments for model selection. \textbf{bold}: best performance in \textcolor{red}{an (aggregator, GraphNorm [+/$-$]) combination}; \textbf{*}: the overall best performance.}
    \TblLab{valid-full-mcf}
\end{table*}

\section{MLAP encompasses JK given same linear aggregator\SecLab{app-b}}

If both an MLAP model and a JK model have linear aggregators with the same form, one can prove that the MLAP model encompasses the JK model.
Here, we consider the relationship between \textit{MLAP-Sum} and \textit{JK-Sum} as an example.

From \EqRef{jk-pool}, a \textit{JK-Sum} model computes the graph representation as
\begin{equation}
    \MlapGraphEmb{\MlapGraph} = \sum_{\MlapNode \in \MlapNodes} \mathrm{softmax} \left( \MlapGateFn (\MlapNodeEmb{\MlapNode}{\mathrm{JK}}) \right) \MlapNodeEmb{\MlapNode}{\mathrm{JK}}.
\end{equation}
Let $a_\MlapNode = \mathrm{softmax} ( \MlapGateFn (\MlapNodeEmb{\MlapNode}{\mathrm{JK}}) )$ be the attention value for each node $\MlapNode$.
Then, from Eqs.~\EqRefBare{jk-hjk} and \EqRefBare{jk-sum},
\begin{equation}
    \MlapGraphEmb{\MlapGraph}
        = \sum_{\MlapNode \in \MlapNodes} a_\MlapNode \sum_{\MlapLayer=1}^{\MlapNumLayer} \MlapNodeEmb{\MlapNode}{\MlapLayer}
        = \sum_{\MlapLayer=1}^{\MlapNumLayer} \sum_{\MlapNode \in \MlapNodes} a_\MlapNode \MlapNodeEmb{\MlapNode}{\MlapLayer}.
    \EqLab{app-jk}
\end{equation}

On the other hand, from Eqs.~\EqRefBare{mlap-hl} and \EqRefBare{mlap-sum}, \textit{MLAP-Sum} computes the graph representation as
\begin{equation}
    \MlapGraphEmb{\MlapGraph} = \sum_{\MlapLayer=1}^\MlapNumLayer \sum_{\MlapNode \in \MlapNodes} \mathrm{softmax} \left( \MlapLayerGateFn{\MlapLayer} (\MlapNodeEmb{\MlapNode}{\MlapLayer}) \right) \MlapNodeEmb{\MlapNode}{\MlapLayer}.
    \EqLab{app-mlap}
\end{equation}
Therefore, when $\mathrm{softmax} ( \MlapLayerGateFn{\MlapLayer} (\MlapNodeEmb{\MlapNode}{\MlapLayer}) )$ for each $\MlapLayer$ takes the exactly same value as $a_\MlapNode$, \EqRef{app-mlap} takes the same form as \EqRef{app-jk}.
It indicates an \textit{MLAP-Sum} model has greater expressivity than a \textit{JK-Sum} model because $\mathrm{softmax} ( \MlapLayerGateFn{\MlapLayer} (\MlapNodeEmb{\MlapNode}{\MlapLayer}) )$ can take different values for each $\MlapLayer$.

We can prove that MLAP encompasses JK using other linear aggregators, \EG \textit{Weighted} or \textit{Concatenation}\footnote{We did not evaluate JK-Weighted nor MLAP-Concatenation in this study.}.
On the other hand, we cannot directly apply the same discussion for nonlinear aggregators like LSTM-Attention.
Still, our experiments showed that MLAP models with linear aggregators tended to perform better than JK with nonlinear aggregators.

\section{Why does MLAP-Weighted perform worse than MLAP-Sum in some datasets?\SecLab{app-c}}

In the synthetic dataset and ogbg-ppa, the \textit{MLAP-Weighted} architecture performed worse than \textit{MLAP-Sum}.
However, intuitively, taking balance across layers using the weight parameters sounds reasonable and effective.
In this appendix section, we show the results of preliminary analyses on the cause of this phenomenon.

\FigRef{results-fractal-weights} shows the weight values in the trained 10-layer models with 30 different random seeds for the synthetic dataset, and \FigRef{results-molhiv-weights} shows the weights in 5-layer ogbg-molhiv models with 30 seeds.
The weight values for the synthetic dataset, where MLAP-Weighted was inferior to MLAP-Sum, had big variances, and the weight distribution covered the ``constant weight'' line (dashed horizontal line; an MLAP-Weighted model is virtually equivalent to an MLAP-Sum model if the weight parameters are equal to this value).
It is expected that the desirable weight for each layer is not largely different from the constant weight.
On the other hand, having such weight parameters with big variance indicates instability during the model training.

In contrast, the weight values for ogbg-molhiv, where MLAP-Weighted performed better than MLAP-Sum, had smaller variances, and the distribution deviated from the constant weight line, particularly in Layers 1 and 5.
It is expected that the desirable weight for those layers is indeed different from the constant weight, and the model might adapt to the balance across layers.

This preliminary analyses suggest that, depending on some properties of datasets, the \textit{MLAP-Weighted} architecture can excel \textit{MLAP-Sum}.
We will continue working on the analyses to identify the suitability of each MLAP aggregator to a certain dataset.

\clearpage

\begin{figure}[t]
    \centering
    \includegraphics{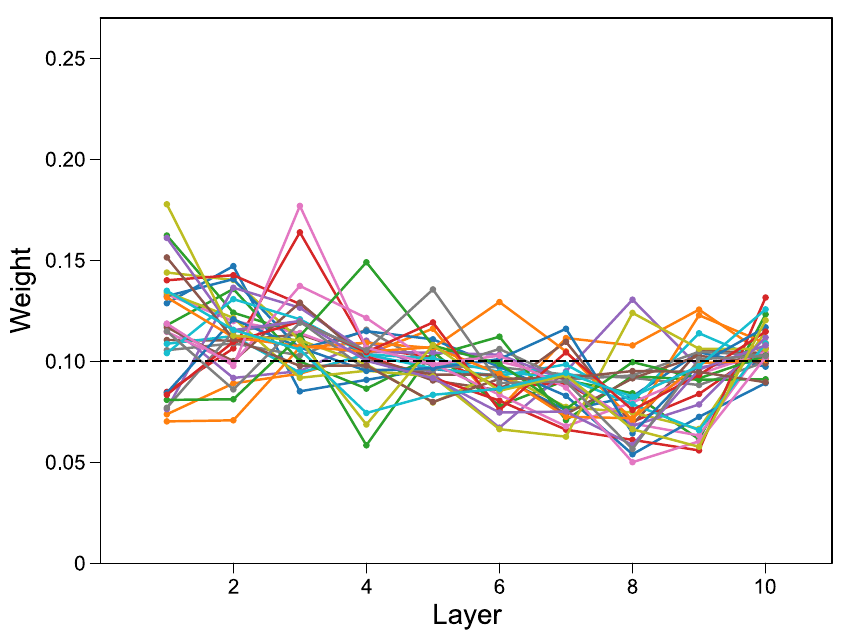}
    \caption{
        Weight parameters of 10-layer MLAP-Weighted models for the synthetic dataset.
        A line shows the weight vector in a model (30 lines in total).
        The dashed horizontal line shows the weight when all layers contribute to the final graph representation equally.
    }
    \FigLab{results-fractal-weights}
\end{figure}

\begin{figure}[t]
    \centering
    \includegraphics{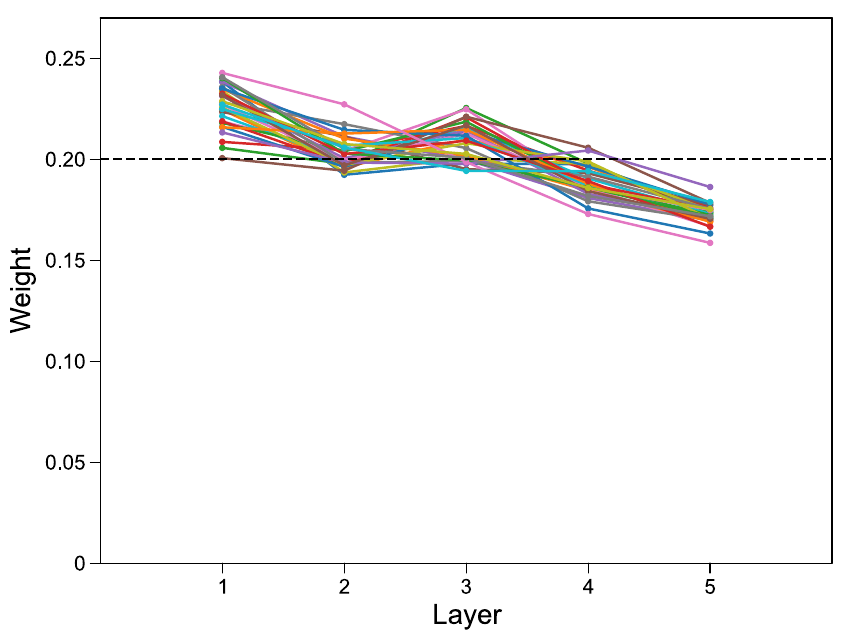}
    \caption{
        Weight parameters of 6-layer MLAP-Weighted models for ogbg-molhiv.
        A line shows the weight vector in a model (30 lines in total).
        The dashed horizontal line shows the weight when all layers contribute to the final graph representation equally.
    }
    \FigLab{results-molhiv-weights}
\end{figure}

\end{document}